\documentclass{article}

\usepackage[preprint]{neurips_2019}
\usepackage[utf8]{inputenc} 
\usepackage[T1]{fontenc}    
\usepackage{hyperref}       
\usepackage{url}            
\usepackage{booktabs}       
\usepackage{amsfonts}       
\usepackage{nicefrac}       
\usepackage{microtype}      
\usepackage{natbib} 
\usepackage{caption} 
\usepackage{float}
\usepackage{graphicx}
\usepackage{subfig}
\usepackage{amsmath}
\usepackage{enumitem}
\usepackage[document]{ragged2e}
\usepackage{multirow}
\usepackage{setspace}
\usepackage{array}
\newcolumntype{P}[1]{>{\centering\arraybackslash}p{#1}}

\title{A Survey of Explainable AI and Proposal for a Discipline of Explanation Engineering}

\author{%
  Clive Gomes \\
  Carnegie Mellon University\\
  Pittsburgh, PA 15213 \\
  \texttt{cliveg@andrew.cmu.edu} \\
  \And  
  Lalitha Natraj \\
  Carnegie Mellon University\\
  Pittsburgh, PA 15213 \\
  \texttt{lnatraj@andrew.cmu.edu} \\
  \AND
  Shijun Liu \\
  Carnegie Mellon University\\
  Pittsburgh, PA 15213 \\
  \texttt{shijunl@andrew.cmu.edu} \\ 
  \And
  Anushka Datta \\
  Carnegie Mellon University\\
  Pittsburgh, PA 15213 \\
  \texttt{anushkad@andrew.cmu.edu} \\
}

\begin{document}

\maketitle

\doublespacing

\begin{abstract}
In this survey paper, we deep dive into the field of Explainable Artificial Intelligence (XAI). After introducing the scope of this paper, we start by discussing what an “explanation” really is. We then move on to discuss some of the existing approaches to XAI and build a taxonomy of the most popular methods. Next, we also look at a few applications of these and other XAI techniques in four primary domains: finance, autonomous driving, healthcare and manufacturing. We end by introducing a promising discipline, “Explanation Engineering,” which includes a systematic approach for designing explainability into AI systems.
  
\end{abstract}

\section{Introduction}

As Artificial Intelligence (AI) technologies are being increasingly used to make vital decisions and perform autonomous tasks, providing explanations that allow users to understand the AI has become a ubiquitous concern in human-AI interaction. However, the manner in which AIs make their decisions is often not explained or is even unexplainable. Thus, attention to explainable AI (XAI) is necessary. 

Trust in computer software results from (1) inspection of code; (2) understanding of program logic; and/or (3) testing.  The first two are nearly impossible in many AI systems, and such systems are so complex that thorough testing cannot be performed, or would be too dangerous to perform (e.g., in autonomous vehicles).  An alternative is to have the AI system explain itself, or to have another AI system examine the first AI system to determine how it works.  Without explainability, it seems unlikely that AIs will be trusted to perform important tasks correctly.  

Explanations not only lead to detection of error, but can also provide avenues for improvement in an AI system.  It also aids in understanding the strengths and weaknesses in AI systems.  In this report, we discuss the nature of explanation, known XAI methods, and their applications in different domains. As a conclusion, we also introduce a promising discipline, Explanation Engineering, which includes a systematic approach for designing explainability into AI systems.


\section{What is an Explanation?}

We start by establishing what an explanation is in the context of Artificial Intelligence. The term has been widely used in recent literature along with similar terms like interpretability, understandability, comprehensibility and transparency, and the perceived definitions seem to overlap between various references. We suspect that this is because there are no universally agreed-upon definitions of these terms and authors assume that the reader understands these terms and the differences between them solely from the context in which they appear. 

In this section, we aim to disambiguate between these terms and provide a thorough discussion of what constitutes an explanation, who the explanation is targeted towards, and why we need explainability in AI.   

\subsection*{Definition}

Explainability is a rather vague term with no formal, widely agreed-upon definition. Yet, authors use it widely and leave it to readers to interpret what a suitable explanation may be for their intended use. It is important to understand that a satisfactory explanation will be different for different scenarios. We cannot derive a general definition of explainability that works for every task in every domain however, we did find the following definition by \cite{arrieta2019explainable} to be most satisfactory among the ones we have seen.

\begin{quote}
\emph{Given an audience, an explainable Artificial Intelligence is one that produces details or reasons to make its functioning clear or easy to understand}
\end{quote}

From the definition, we can identify some factors that should influence the creation of an explanation:

\subsubsection*{(1) Audience}

The audience refers to the group of people towards whom the explanation is targeted. A good explanation should satisfy its audience and be presented in a manner that is suitable for their understanding. However, the “audience” for an AI is not unique. We can identify four major groups as target audiences:

\begin{enumerate}[label=\Alph*.]

\item AI engineers: These are the people responsible for building the model. An explanation that would satisfy AI engineers needs to explain the technicalities of the model so that the engineer understands its operation and can make modifications if required.  

\item Domain specialists: A domain expert is the person who understands the data as part of a domain and may be able to predict outputs given specific inputs. The explanation provided to a domain expert must convince the expert that the model is arriving at the conclusion in a similar manner to what they would expect. 

\item End consumers: The end consumer is the entity that the model directly affects. An explanation targeted towards an end consumer should be more high-level and be presented in a way that garners their trust.

\item Ethicists, lawmakers, insurance companies: Members of this group want an accurate assessment of risk associated with an AI system. For lawmakers, it means that they need to trust the system enough and believe that the advantages outweigh the risks to allow its use by the general public. Insurance companies need to understand risks associated with AI systems to conduct a risk analysis and establish policies for machines utilizing AI. Ethicists are interested in the impact AI has on society as a whole. They are interested in establishing safe interaction between humans and AI, focusing on data privacy, model biases and feedback loops.

\end{enumerate}

For each audience, a different type of explanation may be appropriate. Imagine we have a cancer detection model that detects and classifies the stages of cancer on an X-ray. The domain specialist here would be an oncologist who is trained to look for clumps of cells presenting abnormalities that could be cancerous tumors. The model needs to convince them that it also looks for such cells and applies an effective method of identifying them. 

A good explanation that the model may be able to give is like “these clumps of cells appear lighter than the surrounding areas indicating that the tissue here is of higher density and might be cancerous.” This explanation is suitable for an expert who understands how cancer presents on an X-ray. 

An AI engineer would be more interested in understanding how the model is detecting cancerous regions in an image. A visual explanation such as a heatmap of where the model is focusing after each layer would be a suitable explanation for the engineer, allowing them to understand where the learning is happening. Additionally, a better explanation could be drawn from these heatmaps which explains why those regions are being focused on. This would entail a textual explanation such as “this cluster of pixels looks more like cancerous cells than healthy cells in the given area.” The model could also draw examples from the training set that it thinks look similar to establish why it is giving the result that it is. This would help the AI Engineer improve the model or identify its deficiencies such as overfitting, underfitting, data skew and so on. 

\begin{figure}[H]
    \centering
    \includegraphics[width=\linewidth]{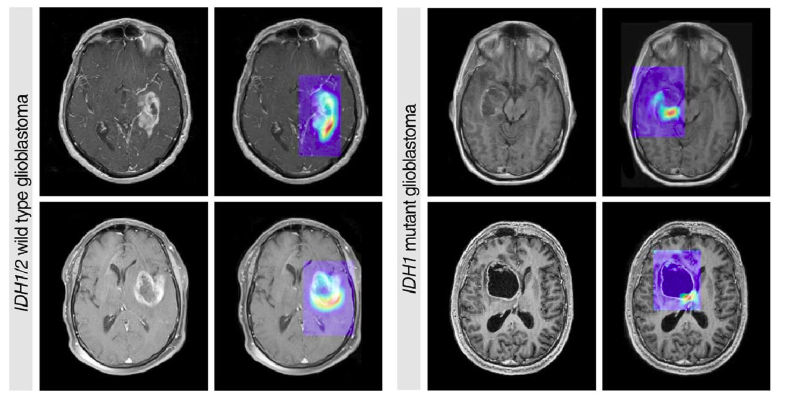}
    \caption{Brain scans that correctly identity the location and type of cancer (\cite{https://doi.org/10.3322/caac.21552})}%
    \label{figure:cancer_type}
\end{figure}

The end user here is the patient, who must feel confident in the diagnosis. The patient does not need to understand the inner workings of the model, nor do they need to understand the science behind it. An explanation suitable for a patient would be one that attempts to show the model’s learning by showing examples in comparison to human experts. A comparison of X-rays labeled by humans and those labeled by the model demonstrating that both give similar results could convince the patient that the model is at least as good as a doctor. For most AI applications, the end user places their trust in the domain expert and convincing the domain expert that the model performs well and gaining their approval and trust can in turn also convince the end user. If some end users are still not convinced, we may provide them with the explanation provided to domain experts. This may happen when a domain expert such as a doctor, who can also be a patient, goes in for an X-ray. The brain scans, shown in Figure \ref{figure:cancer_type}, correctly identify the exact location and type of cancer, verified by human experts (\cite{https://doi.org/10.3322/caac.21552}).

Ethicists, lawmakers and insurance companies will be asked to approve this AI before it can be used in medical diagnosis. These regulatory bodies, such as the U.S. Food and Drug Administration, have procedures in place to test the efficacy and risks of a new medical tool. The AI system must pass their tests to be approved for use. To do so, an in-depth understanding of exactly how the diagnoses are made is needed such that the FDA can assess not only the accuracy, but also patient-centric behavior of the model such as maintaining privacy and eradicating bias. The FDA in particular requires the model to undergo a randomized controlled trial(RCT) where it must perform with an accuracy of over 80\%. Further regulations for AI systems are still in the process of being established.  

\subsubsection*{(2) Truthfulness}

A model’s explanation should accurately reflect the internal workings of the model. The explanation provided by a model should not be different from what is actually happening inside the model. This may happen when a second model is used to explain the main model. An example of this would be an explanation model trying to explain the workings of a self-driving vehicle AI model. Assume we have a vehicle that uses an AI model that gets input from the camera and outputs controls in terms of steering and acceleration commands. Another model tries to explain why the AI model chooses these actions. The explanation model receives the input image, intermediate images and results from the first model and outputs why the first model chooses the action it does such as “the vehicle slows down since there is a pedestrian approaching the road” or “the vehicle turns right since the GPS says so.” 

When training this model, it may learn that some explanations are received better by the user and it may start giving explanations that the user will like even if it does not correspond to what the first model is doing. A user is more likely to be happier with an explanation like “the vehicle stopped because of a red light” than “the vehicle stopped because it started to rain.” This is because the user expects the car to stop at a red light but not when it rains. A faulty AI model may stop every time it rains but the explanation model may convey that it is stopping for a red light which is untruthful and does not help build trust in the model.  

When adding explanations to an AI model, we have a choice to either incorporate the explanation into the model and train the model to give satisfactory results and also satisfactory explanations or we can add an additional observer model that intercepts values from different layers of the original model and builds an explanation from it. Since the explanation is tied into the working of the model for the first method, we expect it to remain truthful. However, when creating an observer model to explain an AI system, we need to be careful about how we train them to avoid receiving untruthful explanations. This can happen if the observer focuses on giving objectively good explanations without understanding the process of the underlying model completely. It may happen accidently or could also be a malicious decision to pass regulation guidelines. Regardless, we should be wary about observer systems and rigorously test them before trusting them. 

\subsubsection*{(3) Personal vs Formal Explanation Evaluation}

A personal explanation is one that cannot be quantitatively measured for adequacy. How good a personal explanation is can be determined by the audience based on whether or not they were satisfied by the explanation. A formal explanation, on the other hand, is one that can be quantitatively rated, which also allows us to compare explanations when necessary. 

Currently, most explanations are personal explanations and are evaluated by methods such as surveys. This allows us to get a general idea of how an explanation may be perceived by an audience but may also let personal biases seep in. Surveys must be carefully designed as to not be suggestive and the answers must be normalized to get a better understanding of how a model is performing. Personal explanations are easier to design since there is no need to add a quantitative measure. If surveys are adequately designed and still satisfy the target audience, then these explanations may be good enough. 

A formal explanation does not depend on surveys but can be quantitatively evaluated for adequacy. Mathematical definitions can be used to rank the quality of the explanation models. Formal explanations are also appropriate when experiments that involve humans cannot be applied for some reason or when the proposed method has not reached a mature enough stage to be evaluated by human users. It also allows for fair comparison between different explanation methods. An existing formal evaluation framework is the PDR, which consists of 3 desiderata that should be used to select interpretation methods for a particular problem: predictive accuracy, descriptive accuracy, and relevancy. These terms are defined as follows:

\begin{enumerate}[label=\Alph*.]
\itemsep0em 
\item Predictive Accuracy: Accuracy on the test set
\item Descriptive Accuracy: The degree to which an interpretation method objectively captures the relationships learned by machine learning models
\item Relevancy: An interpretation is relevant if it provides insight for a particular audience into a chosen domain problem.
\end{enumerate}

The problem we face currently is that there is no agreed upon way to measure descriptive accuracy and relevancy. As such, the idea of having a formal evaluation of explanations is desirable but unachievable.

\subsection*{Explanation vs Observation}

When we talk about explanations, we want the model to provide a deeper understanding of its workings than what we can already observe. We can always observe an AI model by getting the architecture and weights of the system. Given enough time and a set of inputs, we can also manually get an output from the model. This task is more intuitive for some AI techniques than others. Observing a linear regression model or a decision tree is fairly easy while observing a neural network is considerably harder and we do not gather much information from them. Observing simpler models such as a linear regression model may inherently explain aspects of the model such as relative weights of different features towards the final answer. However, we cannot easily or intuitively get the same kind of explanation from other easily observable models such as trees.

\begin{figure}[H]
    \centering
    \includegraphics[width=\linewidth]{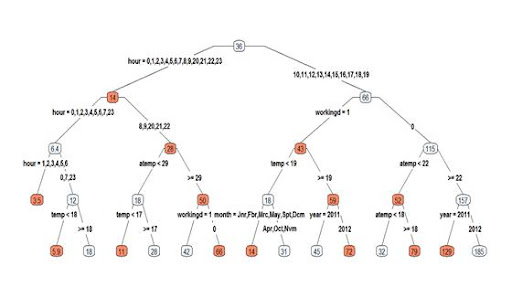}
    \caption{Decision tree explanation from Kang (2015)}%
    \label{figure:dtree}
\end{figure}

\begin{figure}[H]
    \centering
    \includegraphics[width=\linewidth]{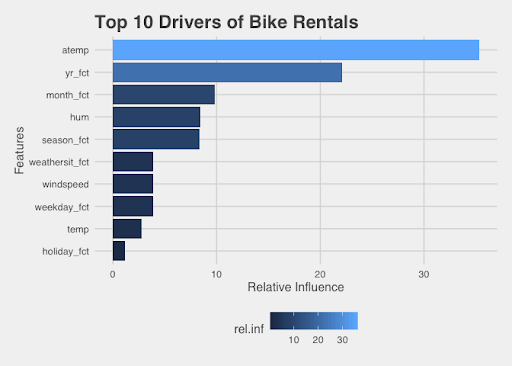}
    \caption{Explanation showing influence of each variable from Spangler (2019)}%
    \label{figure:bikes}
\end{figure}

For example, we can fully observe a decision tree but we also want to understand how each variable plays a part in the final decision; simply looking at a decision tree as presented in Figure \ref{figure:dtree} does not enable us to comprehend which variables are positively or negatively influencing the model’s outcome. This gets even more difficult to do when there is more than one tree as in the case of random forests. An explanation is taking observations and presenting it in a more digestible format that out of which we can get more value. One way to do this is to measure the influence of each variable on the final decision as done in Figure \ref{figure:bikes}. 

\subsection*{Need for Explainability}

Now that we have presented what an explanation is, we touch upon why adding explanations to a project is beneficial despite the additional effort. 

\begin{itemize}
    \item Decisions derived from AI systems ultimately affect human lives. From autonomous vehicles to medical systems, AI is used in systems that can very directly put human lives at risk if the systems are not robust. As such, providing an explanation of what the model is doing is extremely important in getting the system approved for public use and convincing customers of their safety.   
    \item Humans are reluctant to adopt new techniques that are not directly interpretable or tractable. We tend to despise change and AI systems are a huge change from what we are used to. Explanations for these can help the transition and boost user confidence in AI systems.  
    \item Explanations help ensure impartiality in decision making. AI systems for hiring, credit scoring or recidivism prediction, working with sensitive data such as race, gender and age can extract unfair patterns based on these features that occur in historical data. Explaining these AI systems allows us to monitor the use of such sensitive data in decision making, leading to fairer systems going forward.  
    \item Explanations facilitate the provision of robustness. AI systems may behave unpredictably in situations they have not previously experienced. However, if we have explanations of how the AI makes decisions in predictable situations, we may be able to extrapolate to unseen situations as well.
    \item Explanations of AI models also act as an insurance that only meaningful variables infer the output. AI models are prone to overfitting on the training data and adding explanations might provide AI engineers with a sanity check to ensure that the model is only using relevant information from the training data. This is especially important for tasks that require the model to parse signal input such as audio, text or images where some part of the input may contain much more useful information than the others.
    \item There exists a gap between research and business sectors which impedes the full penetration of the newest ML models. This is primarily because AI systems have historically been black box models and any mistakes that they make are a liability for the business employing the system. Explanations attempt to open the black box, building trust between the AI models and businesses looking to employ them.  
    \item There are no/strict regulations for the legal use of AI systems and ethical concerns exist. Regulations for AI models are either extremely strict(such as the ones in the European Union) or simply ban most AI systems. Again, this is because there has been very little effort to explain AI systems. The EU has recently recognised that AI systems can be beneficial but have strict laws that would entail explainability of the model.
    \item Explanations can help with model debugging, monitoring and model audit. Seeing how the system works may help AI Engineers understand where a faulty system is failing. Similarly, after a system is deployed, we can use explanations for monitoring and auditing. 
\end{itemize}

\subsection*{Good Explanations}

A good explanation can add value to the original model and grant it certain attributes that it did not possess before, some of which are as follows:

\begin{itemize}
    \item Trustworthiness - A good explanation may boost confidence of whether a model will act as intended when facing a given problem. It also allows for regular manual intervention if required and encourages regular re-evaluation while in use. Going back to the problem of detecting cancer in X-ray images, a model that gives an explanation like “These clumps of cells appear lighter than the surrounding area indicating that the tissue here is of higher density and might be cancerous” would be perceived as more trustworthy than one that gives an explanation like “The image looks different than a normal, healthy brain” which in turn would be more trustworthy than a black box model.  
    \item Causality - Explainable models might ease the task of finding relationships that, should they occur, could be tested further for a stronger causal link between the involved variables. One example is using an explainable model to learn about causal relationships in objects. A neural network trained to spot cars, explains that a car has four wheels, 4 doors, a hood, bumpers, etc and finding some subset of these indicates that a car is present in the image. This model can be extended to find the spatial relation between different parts of the car and hence find the position or angle of the car. This is establishing causality between these features that have been picked up by the explanation model. The field of Causal AI takes a deeper look into this.
    \item Transferability - Understanding of the inner relations taking place within a model facilitates the ability of a user to reuse this knowledge in another problem. A text to speech model trained on an orthographically transparent language(languages where written form corresponds closely to the spoken form such as Hindi) may not work for an orthographically deep language such as English if the model learns direct grapheme-to-phoneme translations. Lack of proper understanding of the model might drive the user toward incorrect assumptions and they may try to train the same model for English data. However, if we have an explanation for the model, we will know that each grapheme maps to a single phoneme and there is no interaction between graphemes that affects the phoneme(unlike English where individual letters may be pronounced in several ways depending on context). However, knowing this, we can confidently train the same model on several other Indian languages that are also orthographically transparent and expect comparable results.
    \item Informativeness - Explanation of a model lets us extract information about the inner relations of a model. This includes information about where the model is ‘looking’ to get the output for visual, audio and language(eg. Generation, translation) AI models. Explanations can also extract input feature influence. 
    \item Confidence - Understanding how an AI model works allows us to comment on its generalisability and stability. An explanation of a neural network that recognises dogs in images may tell us that the model can detect a dog only if the dog’s face covers at least 10\% of the image. This helps us establish stability of the model and make inferences like the model may not work for collages of dog photos. The explanation will also tell us what it thinks a dog looks like - two ears, four legs, fur arranged in some reasonable configuration. This tells us that the model will likely miss dogs that are partially occluded. It also tells us that other animals may be misclassified as dogs if they match these characteristics well enough. This gives us an idea of the generalisability and specificity of the model.  
    \item Fairness - Biases seep into AI models from historically biased data. Adding explainability into a model would help us observe where these biases are coming from. For example, a hiring AI model is more likely to reject women and accept men even if they are equally qualified. A naive solution would be to remove gender from input features. This however, led to the model picking up on proxy variables and adversely rejecting candidates who graduated from women’s colleges. Such proxy variables may be overlooked unless we try and explain the model and seek them out.
    \item Accessibility - This is the property that allows end users to get more involved in the process of improving and developing a certain ML. For example, Facebook’s ‘Why am I seeing this post/ad?’ feature allows the user to understand why the post or ad was recommended to them. It also allows for transparency as the user is able to see what Facebook knows about them and helps them understand how their actions may influence future recommendations. 
    \item Interactivity - This is the ability of a model to be interactive with the user. Continuing from the previous example, options such as not interested in {user} or not interested in {topic} also allow the user to customize the model to their preferences. 
    \item Privacy awareness - Not being able to understand what has been captured by the model and stored in its internal representation may entail a privacy breach. An AI model usually does not train on sensitive and protected features such as names, gender, address unless required, however this information may slip into the model anyway from other data. For example, a question answering model trained on letters, where addresses are marked out, may still pick up addresses of people or establishments if they appear in the main body of the letter. An explainable model would point to where the model found the answer to “Where does Bill live?” and we could remove that data from the model whereas a black box model will not be able to do so.
\end{itemize}

\section{Explainable AI Techniques}

Different explanations come from a different view or aspect of an AI model. In this section, we focus on different XAI methods and build up a general taxonomy of them. There are two main ways to classify XAI methods: (1) Model Agnostic vs Model Specific methods; and (2) Global vs Local Explanation techniques. Note that these concepts can overlap each other. After exploring these topics, we provide examples of a few Model Agnostic and Local techniques; these methods, and the explanations they result in, may not satisfy everyone, nonetheless, they are popular in the XAI field. At last, we will have a look at model specific methods.

\subsection*{Taxonomy of XAI Concepts}

\subsubsection*{(1) Scope of Interpretability—Global vs Local}

Global interpretability is usually used to describe the understanding of how a model works with inspection of model concepts, while local interpretability usually refers to explaining each individual prediction.

In the context of classifying medical images, local interpretability refers to the ability to ask “which features of this particular image led to it being classified in this particular way?” By global interpretability we mean the ability to ask ‘which common features were generally associated with images assigned to this particular class?’

Let’s take a look at a more complicated example. It seems that global interpretability is a higher standard of understanding an AI model. But it is difficult to directly find a global explanation to a black-box model. Local explanations can be easy to get and it can actually help us reach a global explanation. Below, we provide an example for the same in medical analysis.

\begin{figure}[H]
    \centering
    \includegraphics[width=\linewidth]{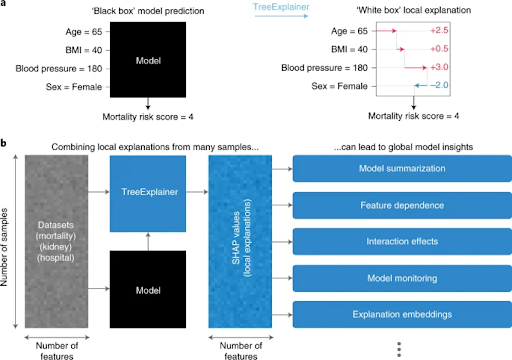}
    \caption{An example of a local explanation using Shapley values}%
    \label{figure:shaply}
\end{figure}

In Figure \ref{figure:shaply}.a, we see an example of a local explanation based on assigning a numeric measure of credit to each input feature with a local method called Shapley Values. Figure \ref{figure:shaply}.b combines many local explanations to represent a global structure while retaining local faithfulness to the original model. Computing local explanations across all samples in a dataset enables development of many tools for understanding global model structure. What this implies is that for this “Age=65, BMI=40, Blood pressure=180, Sex=Female” as the input, we get a local explanation for the contribution of these for number like +2.5, -2.0 and their category respectively; a positive number implies a positive effective for mortality risk score while a negative one indicates the opposite. The absolute value simply reflects how big the effect of the item’s contribution is.

Although this can be different for every individual in the dataset, a considerable number of local explanations can be helpful to obtain global insights. For example, the average of the “contribution score” can reflect the overall feature dependence and importance of these features. The range in which these features lie can also be a good insight for model summarization. 

\subsubsection*{(2) Model-Agnostic vs Model-Specific Techniques}

Model-specific interpretation techniques are specific to a single model or group of models.  These techniques depend heavily on the architecture and capabilities of a specific model. In contrast, model-agnostic tools can be used on any machine learning model, regardless of how complicated they are. These agnostic methods usually work by analyzing input features and output pairs.

To better explain what model-agnostic techniques are, we use the explanation of a classification task where YouTube comments are marked as spam or normal using a technique called LIME.

The black box model is a deep decision tree trained on the document word matrix. Each comment is one document (= one row) and each column is the number of occurrences of a given word. Short decision trees are easy to understand, but in this case the tree is very deep. Also in place of this tree there could have been a recurrent neural network or a support vector machine trained on word embeddings (abstract vectors). Let us look at the two comments of this dataset and the corresponding classes in Figure \ref{figure:youtube1} (1 for spam, 0 for normal comment, spam comments are typically characterized by the presence of replies that are irrelevant to the blog entry, along with a link that leads to the commenter’s website):

\begin{figure}[H]
    \centering
    \includegraphics[width=\linewidth]{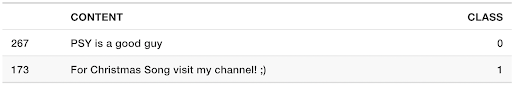}
    \caption{Example of a YouTube comment and class assigned}%
    \label{figure:youtube1}
\end{figure}

\begin{figure}[H]
    \centering
    \includegraphics[width=\linewidth]{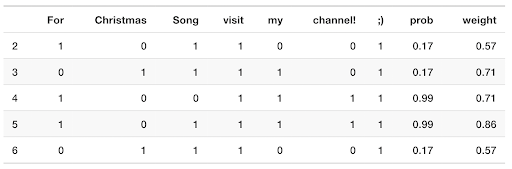}
    \caption{Example of the contribution of each word of a YouTube comment to a class}%
    \label{figure:youtube2}
\end{figure}

Each column in Fig. \ref{figure:youtube2} corresponds to one word in the YouTube comment. Each row is a variation of the original word counts: a 1 means that the word is part of this variation and 0 means that the word has been removed. This allows us to eliminate certain words from the comment and observe its effect on the output. The "prob" column shows the predicted probability of the comment being spam while the "weight" column shows the proximity of the variation to the original sentence, calculated as 1 minus the proportion of words that were removed, for example if 1 out of 7 words was removed, the proximity is 1 - 1/7 = 0.86.

Figure \ref{figure:youtube_table} shows the estimated local weights found using the LIME algorithm:

\begin{figure}[H]
    \centering
    \includegraphics[width=\linewidth]{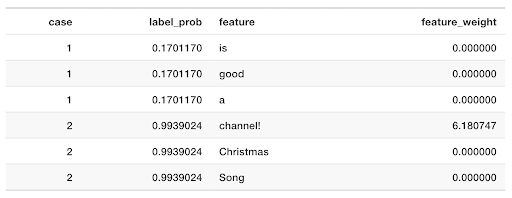}
    \caption{Contribution of each feature to the label assigned}%
    \label{figure:youtube_table}
\end{figure}

The word "channel!" indicates a high probability of spam. For the non-spam comment, no non-zero weight was estimated because, regardless of which word is removed, the predicted class remained the same.

\subsection*{Examples of XAI Methods}

\subsubsection*{(1) Example-Based Explanations}

Example-based explanations help humans construct mental models of the machine learning model and the data the machine learning model has been trained on. It especially helps people understand complex data distributions. But what are example-based explanations? We often use them in our jobs and daily lives. Let us start with some examples.

\begin{quote}
    “A physician sees a patient with an unusual cough and a mild fever. The patient's symptoms remind her of another patient she had years ago with similar symptoms. She suspects that her current patient could have the same disease and she takes a blood sample to test for this specific disease.”
\end{quote}

\begin{quote}
    “A data scientist works on a new project for one of his clients: analysis of the risk factors that lead to the failure of production machines for keyboards. The data scientist remembers a similar project he worked on and reuses parts of the code from the old project because he thinks the client wants the same analysis.”
\end{quote}

These stories illustrate how we humans think in terms of examples or analogies. The blueprint of example-based explanations is as follows: thing B is similar to thing A and A caused Y, so we predict that B will cause Y as well. 

Implicitly, some machine learning approaches are example-based. Decision trees partition the data into nodes based on the similarities of the data points in the features that are important for predicting the target. A decision tree gets the prediction for a new data instance by finding the instances that are similar (in the same terminal node) and returning the average of the outcomes of those instances as the prediction. The K-Nearest Neighbors (KNN) method also works explicitly with example-based predictions.

Example-based explanations are mostly model-agnostic, because they make any machine learning model more interpretable. The difference between example-based and model-agnostic methods is that the example-based methods explain a model by selecting instances of the dataset and not by creating summaries of features. Example-based explanations also only make sense if we can represent an instance of the data in a humanly understandable way. This works well for images, because we can view them directly. For a new instance, a knn model locates the K-Nearest Neighbors (for e.g. the K=3 closest instances) and returns the average of the outcomes of those neighbors as a prediction.

\subsubsection*{(2) LIME}

LIME is short for Local Interpretable Model-Agnostic Explanations. Each part of the name reflects something that we desire in explanations. Here, “local” refers to local fidelity—we want the explanation to reflect the behavior of the classifier "around" the instance being predicted. This explanation is useless unless it is interpretable, i.e., a human must be able to make sense of it. LIME is able to explain any model without needing to 'peek' into it and, therefore, it is model-agnostic.

\vspace{3mm}

\emph{\textbf{Core Thoughts of LIME: Is the reason acceptable?}}

\vspace{2mm}

Let’s take a look at an example of a text classification task.

\begin{figure}[H]
    \centering
    \includegraphics[width=\linewidth]{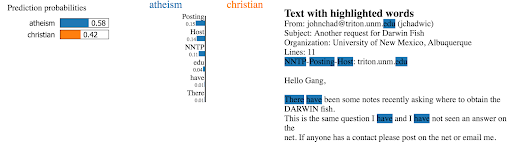}
    \caption{Example of a text classification task using LIME}%
    \label{figure:lime}
\end{figure}

In Figure \ref{figure:lime}, the famous 20 newsgroup dataset was used to classify news as related to “Christian” or “Atheism.” Training a random forest with 500 trees, a test set accuracy of 92.4\% was achieved, which is surprisingly high. If accuracy was our only measure of trust, we would definitely trust this algorithm. What LIME does is remove and put back words in the text, one by one, to see the change in probabilities. “Posting: 0.15” means that if we remove “Posting” from the text on the right, the probability of atheism will reduce to 0.58 - 0.15 = 0.43, and the probability of christian will increase by 0.15.

A little further exploration shows us that the word "Posting" (part of the email header) appears in 21.6\% of the examples in the training set but only two times in the class “Christian.” This is also the case for the test set, where it appears in almost 20\% of the examples and twice in 'Christian'. This kind of quirk in the dataset makes the problem much easier than it is in the real world, where this classifier would not be able to distinguish between Christian and Atheism-related documents. This is hard to see just by looking at accuracy or raw data, but easy once explanations are provided.

\vspace{3mm}

\emph{\textbf{Inspiration behind LIME}}

\vspace{2mm}

\begin{figure}[H]
    \centering
    \includegraphics[width=\linewidth]{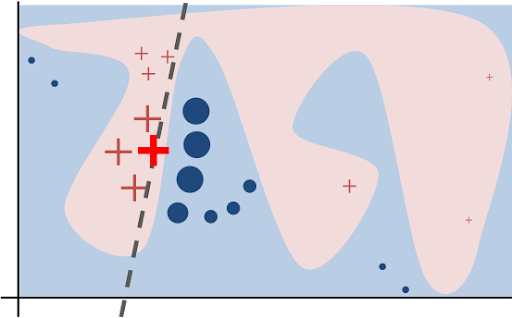}
    \caption{Figure to illustrate the intuition behind LIME’s decision function}%
    \label{figure:lime_df}
\end{figure}

Intuitively, an explanation is a local linear approximation of the model's behavior. While the model may be very complex globally, it is easier to approximate it around the vicinity of a particular instance. While treating the model as a black box, we perturb the instance we want to explain and learn a sparse linear model around it, as an explanation. Figure \ref{figure:lime_df} illustrates the intuition for this procedure. The model's decision function is represented by the blue/pink background, and is clearly nonlinear. The bright red cross is the instance being explained—let's call it X. We sample instances around X, and weight them according to their proximity to X (weight here is indicated by size). We then learn a linear model (dashed line) that approximates the model well in the vicinity of X, but not necessarily globally.

\subsubsection*{Shapley Values}

A prediction can be explained by assuming that each feature value of the instance is a "player" in a game where the prediction is the payout. Shapley values—a method from coalitional game theory—is a model-agnostic technique that tells us how to fairly distribute the "payout" among the features.

Consider the following scenario:

\begin{quote}
    “You have trained a machine learning model to predict apartment prices. For a certain apartment it predicts €300,000 and you need to explain this prediction. The apartment has an area of 50 m2, is located on the 2nd floor, has a park nearby and cats are banned.”
\end{quote}

\begin{figure}[H]
    \centering
    \includegraphics[width=\linewidth]{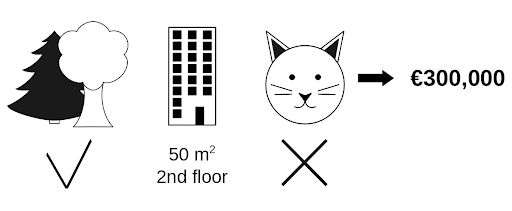}
    \caption{Example for Shapley values}%
    \label{figure:shapley_ex}
\end{figure}

The average prediction for all apartments is €310,000. How much has each feature value contributed to the prediction compared to the average prediction?

The answer is simple for linear regression models. The effect of each feature is the weight of the feature multiplied by the feature value. This only works because of the linearity of the model. For more complex models, we need a different solution. For example, LIME suggests local models to estimate effects. Another solution comes from cooperative game theory: the Shapley value (coined by Shapley) is a method for assigning payouts to players depending on their contribution to the total payout. Players cooperate in a coalition and receive a certain profit from this cooperation.

A game theory problem can be divided into three parts: players, game and payout. The "game" is the prediction task for a single instance of the dataset. The "gain" is the actual prediction for this instance minus the average prediction for all instances. The "players" are the feature values of the instance that collaborate to receive the gain (i.e., predict a certain value). In our apartment example, the feature values park-nearby, cat-banned, area-50 and floor-2nd worked together to achieve the prediction of €300,000. Our goal is to explain the difference between the actual prediction (€300,000) and the average prediction (€310,000): a difference of - €10,000.

The answer could be: the “park=nearby” contributed €30,000; “area=50” contributed €10,000; “floor=2nd” contributed €0; “cat=banned” contributed -€50,000. The contributions add up to -€10,000, the final prediction minus the average predicted apartment price.

How do we reach these contributions?

In Fig. \ref{figure:shapley_ex}, we evaluate the contribution of the cat-banned feature value when it is added to a coalition of “park=nearby” and “area=50.” We simulate the environment such that only park-nearby, “cat=banned” and “area=50” are in a coalition by randomly drawing another apartment from the data and using its value for the floor feature. The value “floor=2nd” was replaced by the randomly drawn “floor=1st.” Then we predict the price of the apartment with this combination (€310,000). In a second step, we remove cat-banned from the coalition by replacing it with a random value of the cat allowed/banned feature from the randomly drawn apartment. In the example, it was “cat=allowed,” but it could have been “cat-banned” again. We predict the apartment price for the coalition of “park=nearby” and “area=50” (€320,000). The contribution of “cat=banned” was €310,000 - €320,000 = - €10,000. This estimate depends on the values of the randomly drawn apartment that served as a "donor" for the cat and floor feature values. We will get better estimates if we repeat this sampling step and average the contributions.

\begin{figure}[H]
    \centering
    \includegraphics[width=\linewidth]{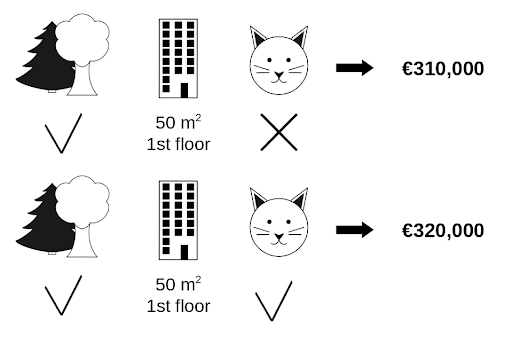}
    \caption{Example for Shapley values}%
    \label{figure:shapley_ex2}
\end{figure}

We repeat this computation for all possible coalitions. The Shapley value is then the average of all the marginal contributions to all possible coalitions. One problem with this method is that the computation time increases exponentially with the number of features. A simple solution for keeping the computation time manageable is to compute contributions for only a few samples of the possible coalitions.

Fig. \ref{figure:shapley_ex2} shows all coalitions of feature values that are needed to determine the Shapley value for “cat=banned.” The first row shows the coalition without any feature values. The second, third and fourth rows show different coalitions with increasing coalition size, separated by "|". All in all, the following coalitions are possible:

\begin{itemize}
    \item No feature values
    \item park=nearby
    \item area=50
    \item floor=2nd
    \item park=nearby + area=50
    \item park=nearby + floor=2nd
    \item area=50 + floor=2nd
    \item park=nearby + area=50 + floor=2nd
\end{itemize}

For each of these coalitions we compute the predicted apartment price with and without the feature value cat-banned and take the difference to get the marginal contribution. The Shapley value is the (weighted) average of marginal contributions. We replace the feature values of features that are not in a coalition with random feature values from the apartment dataset to get a prediction from the machine learning model.

\subsubsection*{(4) Simple Regression Models (Linear and Logistic Regression)}

In this section, we use the example of a linear regression model that predicts the number of rented bikes on a particular day, given weather and calendar information. For interpreting the model’s decision-making process, we examine the estimated regression weights; features consist of numerical and categorical features.

\textbf{Interpretation of numerical features (“temperature”):} An increase in temperature by 1 degree Celsius increases the predicted number of bicycles by 110.7, when all other features remain fixed.

\textbf{Interpretation of categorical features ("weathersit"):} The estimated number of bicycles is -1901.5 lower when it is raining, snowing or stormy, compared to good weather—again, assuming that all other features do not change. When the weather is misty, the predicted number of bicycles is -379.4 lower compared to good weather, given all other features remain the same.

\textbf{More visual explanations:} The information of the weight table (weight and variance estimates) can be visualized in a weight plot. The plot in Fig 3.10 shows the results from the previous linear regression model.

\begin{figure}[H]
    \centering
    \includegraphics[width=12cm]{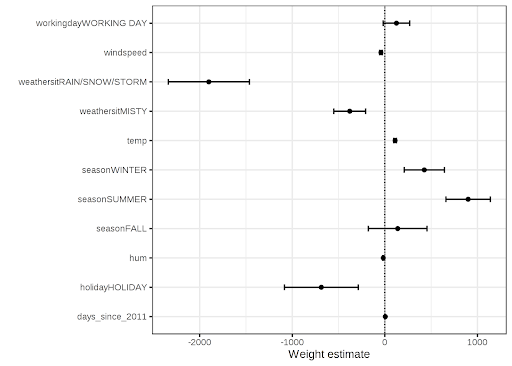}
    \caption{Plot of each weight along with their confidence intervals}%
    \label{figure:regression1}
\end{figure}

\begin{figure}[H]
    \centering
    \includegraphics[width=12cm]{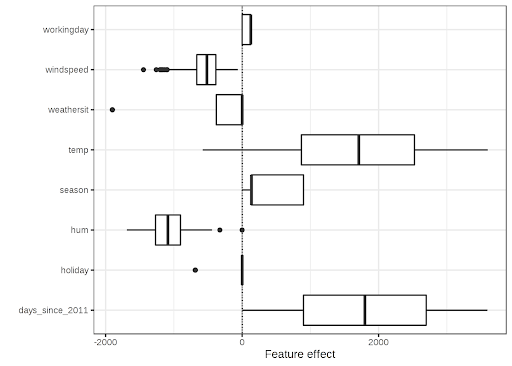}
    \caption{Plot of the product of the input feature and respective weight for all training instances}%
    \label{figure:regression2}
\end{figure}

\subsection*{Summary of XAI Terminology}

Comment on these models: pros and cons, do they satisfy different people’s needs etc.

Some thoughts: Explaining models are easy and understandable, but they are too naive and only applied to elementary models.

LIME is good and acceptable by most users. But it still fails to satisfy some specific users asking questions like “Why does the model think this word means Christian?”

\begin{table}[h] 
  \caption{Summary of XAI techniques}
  \label{table:retrival}
  \raggedleft
  \begin{tabular}{|p{2.5cm}|p{3cm}|p{3.3cm}|p{3.3cm}|}
    \toprule
    \cmidrule(r){1-4}
    XAI methods & Characteristics & Pros & Cons \\
    \midrule	
    LIME & Local, \newline Model Agnostic & Can be applied to any model. \vspace{3mm} \newline Easy to implement. \vspace{3mm}\newline  Does not have rigid requirements on the quality of datasets. & Explanations can be superficial and not convincing. \vspace{3mm} \newline E.g: Fails to answer “Why does the model think this word means Christian?” \\
    \midrule
    Shapley Values & Local, \newline Model Agnostic & Can be applied to any model. \vspace{3mm} \newline Explanations are more convincing than LIME. & Need high-quality datasets. \vspace{3mm} \newline Less available fit cases compared to LIME. \\
    \midrule
    Naive Regression Explanation & Global, \newline Model Specific & Clear, easy and \newline almost the most \newline correct explanations. & Can only be applied to a single set of AI models. \\
    \bottomrule
  \end{tabular}
\end{table}

\section{XAI in Various Domains} \label{sec:data}

Artificial Intelligence has become increasingly popular as a powerful tool to solve difficult problems in multiple domains. The extent to which an AI system needs to be explainable depends on the domain in which it is being used as well as the specific task being performed. This is because some domains and tasks therein require a higher level of transparency in their decision making than others do.

For example, an AI system that recommends movies to users on an online streaming platform would not require as exhaustive an explanation as that of an AI system that predicts whether a patient has a benign or malignant tumor. This is because the nature of the explanation provided depends on the task that the AI system is performing, the audience to which the explanation is being provided, and the consequences of an erroneous recommendation. As a result, different domains and different tasks within each domain require different types of explanations and specific XAI techniques to produce these explanations that are considered satisfactory.

In this section, we describe the latest developments of explainable AI in four selected domains: healthcare, finance, autonomous driving and manufacturing. For each of these domains, we explore the need for an explanation in the domain, AI systems that are typically used for different tasks in this domain, and the latest developments in the XAI techniques used. Furthermore, the subjectivity of the audience within each domain will be taken into consideration when establishing what constitutes a satisfactory explanation. Verification methods to establish the correctness of the explanations produced will also be discussed.

\subsection*{Finance}

The finance domain consists of multiple sub fields including corporate finance, investment banking, financial economics, financial markets, international finance, personal finance and public finance. Since A.I. systems are most widely used for applications such as stock market prediction and credit risk management, this section will be focusing on the sub fields of financial markets and investment banking.

Decisions a company makes with regard to a customer needs to be explained as per the tight regulations imposed within these subfields. For example, the Equal Credit Opportunity Act (\cite{usact}) “requires creditors to provide applicants, upon request, with the reasons underlying decisions to deny credit.” Therefore, an AI model that is deployed in this domain needs to be transparent and explainable. Furthermore, a lack of explainability in an AI model makes it difficult for accountants to conduct a thorough audit of the company’s financial workflow. This is because the level of skepticism that can be applied during the audit will be bottlenecked by the level of explainability exhibited by the AI system.

The audiences for an explainable AI system depend on the task that is being performed. Generally, the audiences include stakeholders that are affected by the decision made by the AI system. This typically includes the company, its customers, as well as the regulatory bodies that require the decisions to be explained.

In the following subsections, three different XAI systems developed for the finance domain will be discussed. The first XAI system provides explanations for deep learning predictions for the task of stock market prediction. The second explains machine learning predictions of risk in credit risk management. Finally, the third looks at stock price prediction using sentiment analysis conducted on financial news articles. Each of these systems will be assessed to determine whether the explanations generated are satisfactory, taking into consideration the specific task being performed.

\subsubsection*{(1) CLEAR (Class-Enhanced Attentive Response) approach for producing visual explanations of deep learning-driven stock market predictions}

The paper by \cite{kumar2017opening} describes an XAI system that explains predictions made by neural networks for the task of predicting stock market movement. The input to the system is a time series of historical trade information regarding a particular stock market index (30 days worth of open, close, highs, lows and trade volumes). This information is presented in the form of a graph so that image processing techniques can be applied. The output of the model is a binary output that indicates whether a stock goes up or down, given the trend of the graph until that time step and the model that performs this classification is a convolutional neural network. 

For the XAI system that explains the binary classification output by this model, the CLEAR approach is used. This involves the generation of a heat map on the input image data (the graph in this case) that aims to explain the decision produced by the system. In this case, the  generated heat map aims to

\begin{enumerate}[label=\Alph*.]

\item Highlight the attentive time windows in the graph that were responsible for the decision made by the deep learning model

\item Depict the attentive levels at each of these attentive time windows (this will represent the level of significance this region had over the prediction made)

\item Depict the class that is associated with each of these highlighted time windows (in this example, the class would be the stock going up, or the stock going down)

\end{enumerate}

\begin{figure}[H]
    \centering
    \includegraphics[width=\linewidth]{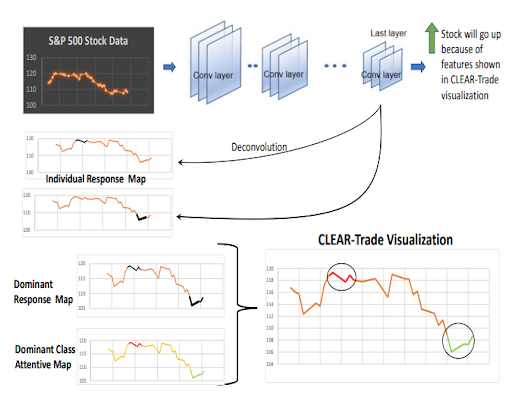}
    \caption{Working of CLEAR Trade as described in \cite{kumar2017opening}}%
    \label{figure:clean}
\end{figure}

The CLEAR algorithm for generating a heat map from the convolutional neural network works as follows:

\begin{enumerate}
    \item From the last CNN layer of the network, the output activations are back projected with a deconvolution operation to compute an individual attentive response map for each kernel associated with a class. 

    \item From these attentive response maps, 2 maps are computed, namely the dominant attentive response map and the dominant state attentive map. The dominant attentive response map shows which parts of the graph influenced the decision of the system most significantly. The dominant state attentive map shows which parts of the graph are associated with which class. 

    \item These two maps are combined into a single map that depicts the final CLEAR-Trade visualization as shown in Fig. \ref{figure:clean}. 
\end{enumerate}

Although this model is able to explain the stock market prediction made by the system effectively through a visual approach, it does not consider any external factors that may affect the prediction that does not rely on the input data provided. This becomes an issue and a potential avenue for improvement because stock movement cannot be predicted from time-series analysis alone but must take into account extrinsic information. Therefore, although the explanation of the prediction may be satisfactory to the audience (which in this case would be a financial analyst), the actual prediction itself is questionable. 

\subsubsection*{(2) Explainable Machine Learning in Credit Risk Management}

\cite{Bussmann2020} have developed an XAI system to explain the task performed by a ML model designed to measure the risks that arise when credit is borrowed employing peer to peer lending platforms by estimating the probability of default of each company. The XAI system aims to explain the resulting probability in terms of the magnitude of impact of each individual explanatory variable on the final outcome. The input to the system includes variables describing the borrowing companies that were used in training the ML model. This is done using Shapley value decompositions. 

The Shapley values of the explanatory variables exhibit the following properties:

\begin{enumerate}[label=\Alph*.]
    \itemsep0em
    \item Local Accuracy: A Shapley value is a unique quantity that can locally linearly approximate the original model output, for a specific input x. 
    \item Missingness: If a feature is locally zero, the Shapley value is also zero
    \item Consistency: If in a second model the contribution of a feature is higher, so will be its Shapley value.
\end{enumerate}

\begin{figure}[H]
    \centering
    \includegraphics[width=\linewidth]{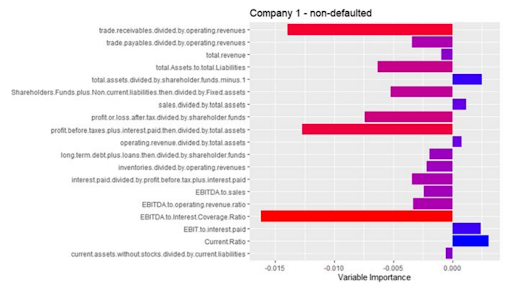}
    \caption{Plot that depicts the contribution of each explanatory variable for Shapley values}%
    \label{figure:company_shapley}
\end{figure}

The graph in Figure \ref{figure:company_shapley} depicts the contribution of each explanatory variable (in terms of its Shapely value) to the prediction made by the ML system on the probability that the company has defaulted. The more red the color, the higher the negative importance. Similarly, the more blue the color, the higher the positive importance.

Contribution of each explanatory variable to the prediction regarding a particular company is provided. However, the explanations stop at “which” variables influence the decision and do not mention “why” those variables influence the decision. Such explanations may not be satisfactory to the audience of financial professionals that analyze the credit risk associated with companies because they would require more insight into why the model interprets some features to have a greater impact than others. 

\subsubsection*{(3) Explainable stock prices prediction from financial news articles using sentiment analysis}

In the paper by \cite{articleGite}, we see another example for the prediction of stock prices using machine learning. The input to the ML model is a preprocessed version of the News Headlines Dataset and a preprocessed Yahoo Finance dataset. An LSTM model is used to perform the stock market prediction. The XAI system developed uses the Local Interpretable Model-Agnostic Explanations (or LIME) technique. LIME is model agnostic in the sense that it can give explanations for any given supervised learning model (as we discussed earlier in this paper). 

When given a prediction model and a test sample, the following steps are taken when using the LIME technique:

\begin{enumerate}
    \item Sampling and Obtaining a Surrogate Dataset: A number of samples in the vicinity of the feature vector being explained are produced following a normal distribution. This sampled set is called the surrogate dataset. Then it produces the output for these samples using the ML model that is being explained. 
    \item Feature Selection: Each row in the surrogate dataset is weighed with respect to how close they are from the original feature vector and its output. Following this, Lasso regularization is used to obtain the most influential features to the decision made by the ML model. 

\end{enumerate}

\begin{figure}[H]
    \centering
    \includegraphics[width=\linewidth]{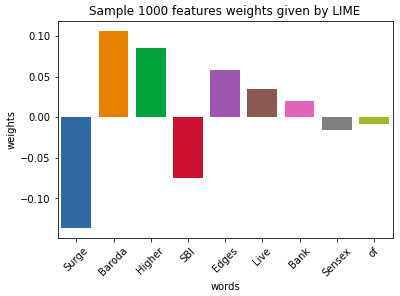}
    \caption{Visual representation of contribution of each feature to the model decision using LIME}%
    \label{figure:company_lime}
\end{figure}

These top features can be visually represented to provide explanation as can be seen in Figure \ref{figure:company_lime}. A negative weight corresponds to a more negative context as can be seen for the word ‘surge’ that has a high negative weight and, therefore, high negative context. 

A final point worth mentioning is that this example takes the ‘The Efficient Market Hypothesis’ into account which states that asset prices cannot entirely depend on obsolete information and market prices react to new information, for example, financial news articles, social media blogs. This implies that the model takes a more holistic approach to its prediction, in contrast to the aforementioned CLEAR approach for the same task of stock market prediction. As a result, the explanation produced by this model gives more information about the reasoning behind the model’s prediction to the audience.

\subsection*{Autonomous Driving}

In the domain of autonomous driving, XAI systems are required primarily to ensure proper functioning of the AI model in the design and testing phase. In addition to this, explanations provide valuable insight into the working of the model. As a result, these explanations can help an AI practitioner understand where the model is failing and identify areas and strategies to improve the existing model. Autonomous driving in particular relies heavily on user trust. If an AI system designed for autonomous driving can provide explanations for its decisions, the trust that the user has towards the system will improve. Finally, XAI systems are required to explain the decisions made by an autonomous car (such as detecting a pedestrian) to answer potential legal, and even ethical, questions (what caused the car to fail in detecting the pedestrian?).

The audiences for an XAI system in the domain autonomous driving can be divided into three categories: the AI practitioners, the users and the regulatory bodies. The goals of an XAI system include system monitoring and improvement for the AI practitioners, trust and safety for users, and, finally, accountability and transparency for the regulatory bodies.

In the following sections, we discuss interpretable learning techniques for self-driving cars from two main papers \cite{kim2017interpretable} and \cite{kim2018textual}. \cite{kim2017interpretable} discusses an XAI system that provides explanations for CNN encoder decoder models using heatmaps. \cite{kim2018textual} generates heat maps as well as textual explanations given a stream of input images. For each of these examples, we will assess whether the explanations produced are satisfactory keeping in mind the audience to which the explanation is targeted.

\subsubsection*{(1) Interpretable Learning for Self-Driving Cars by Visualizing Causal Attention}

\begin{figure}[H]
    \centering
    \includegraphics[width=\linewidth]{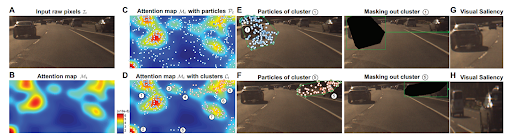}
    \caption{Fig. 4.4: A- Raw input Pixels, B- Attention map that is derived from the coarse grained explanation, C- Randomly sample N points or particles from the attention map, D- Group the samples points into clusters, E and F- Produce a convex hull for each of these clusters and mask out the convex hull in the original input. This masked input is used to see causal effects on predictive accuracy of the model, G and H - Warped visual saliencies for clusters 1 and 5}
    \label{figure:av_map}
\end{figure}

In this paper by \cite{kim2017interpretable}, an approach to producing visual explanations of the decision of steering direction has been described. The first step involves the use of a visual attention model to train a Seq2Seq model (with a CNN encoder) to predict steering angle from an image. The process of attaining the visual explanations for this model are two fold:

\begin{enumerate}
    \item Coarse-Grained Explanations: The attention map that is used in the attention model of the sequence-to-sequence network can be visualized to highlight image regions that potentially influence the network’s output. Some of these are true influences, but some are spurious which elicits the need for a filtering step.
    \item Fine-Grained Explanations: Points on the attention map are clustered to find local visual saliency. This is a causal-filtering step that determines which input regions actually influence the network’s output. The working of the fine-grained explanation can be explained with Figure \ref{figure:av_map}. 

\end{enumerate}

\begin{figure}[H]
    \centering
    \includegraphics[width=\linewidth]{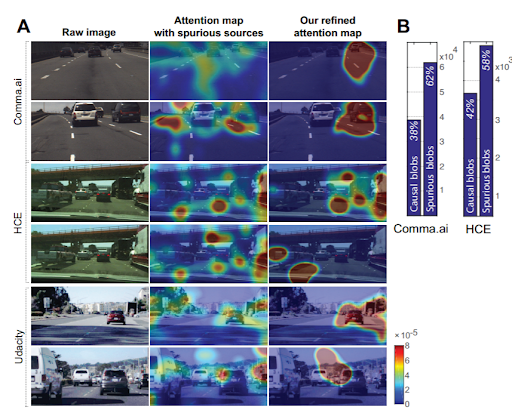}
    \caption{Attention maps generated for the input images}%
    \label{figure:av_gen}
\end{figure}

Figure \ref{figure:av_gen} shows the working of the explanation model on the dataset obtained from the Hyundai Center of Excellence in Integrated Vehicle Safety Systems and Control (HCE) as well as the Comma.ai datasets.

A heat map may not always be a satisfactory explanation, especially when the audience is an AI practitioner who is developing the autonomous driving system for predicting steering direction. This is because the heat map only shows ‘what’ features are important but not ‘why’ the model considers them important. Therefore, a better choice of XAI technique would be one that provides reasoning as to how the decision was made using the important features rather than just identifying the most influential features.

\subsubsection*{(2) Textual Explanations for Self-Driving Vehicles}

\cite{kim2018textual} have developed a system that generates textual explanations for the output of a neural network model that predicts vehicle control commands (acceleration and change of course). 

First, a convolutional feature encoder encodes an input image into a number of feature vectors, each representing a different spatial region in the input image. A stream of such encoded images is then sent through a number of learnable attention maps. These attention maps encode the spatial importance of the objects in the image. The context vectors produced by these attention maps are sent to a LSTM network. The LSTM network outputs the probability for the two vehicle controls acceleration and change of course and, using the labeled training data, the attention maps for this stage are learnt. 

The attention map used by the vehicle controller (the neural network that needs to be explained) represents the regions in the input image that influence the network’s decision. The textual explanation generator must explain the evidence available in the attention map. For example, if the vehicle controller predicts that the vehicle should ‘slow down’ upon detecting a red stop light, the textual explanation must include this evidence as “slow down because of the stop light ahead.” The vehicle controller and the textual explanation generator should, therefore, focus on the same regions of the attention plot. Two methods are discussed in the paper to enforce this alignment:

\begin{enumerate}
    \item Strongly aligned attention where both the vehicle controller and the textual explanation generator use the same attention map.
    \item Weakly aligned attention where the textual explanation uses its own spatial attention map. The Kullback-Leibler (KL) divergence between the attention map of the vehicle controller attention map and the textual explanation generator attention map enforces that the explanations refer to salient features identified by the vehicle controller. 
\end{enumerate}

The actual text generation model works as follows. For strongly aligned attention, the input is the context vectors generated by the vehicle controller’s attention maps. For weakly aligned attention, the input is the context vectors generated by the text generator’s own attention maps. These maps are generated from the image and learnt using the KL alignment loss with respect to the vehicle controller’s attention map. In a variation of the model that produces rationalizations, this loss is not enforced. As a result, the rationalization model does not have any constraints on attention alignment. In addition to these context vectors, the prediction made by the vehicle controller is concatenated with the context vectors and sent as input to an LSTM network. This network then produces the text output at each time instance, generating the probabilities for each word in a predefined dictionary. 

\begin{figure}[H]
    \centering
    \includegraphics[width=\linewidth]{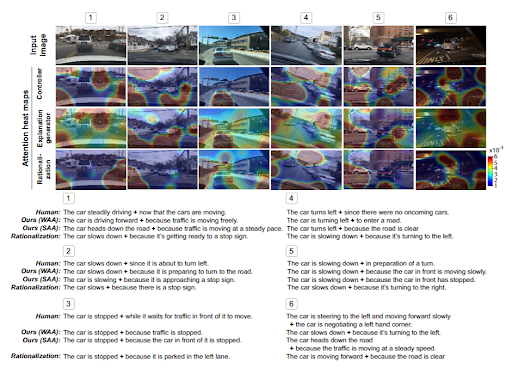}
    \caption{Results of the textual explanations generated}%
    \label{figure:av_textgen}
\end{figure}

The network is trained on text that is divided into two parts—the description and the explanation—separated by a marker <sep>. An example of this would be ‘The car is speeding up <sep> because of a green light’; in this case, ‘The car is speeding up’ is the description and ‘because of a green light’ is the explanation. The network, therefore, learns to output similar text that has both descriptions and explanations. The results can be seen in Figure \ref{figure:av_textgen}. 

This XAI system provides a heat map as well as a rationalization behind the decision taken by the AI in the form of textual explanations. This is a better approach than the previous example because it not only explains the ‘what’ (what parts of the image influenced the decision) but also the ‘why’ (why did this part of the image influence the decision). For example the ‘what’ can refer to the fact that the upper right corner of the image is significant, but the ‘why’ can tell the audience that the green light (that is in the upper right corner of the image) is significant. In addition, the explanations are easy to understand as they are presented in natural language and not abstruse mathematical formulae and/or graphs. This makes these explanations suitable for audiences such as the end users and regulatory authorities who do not need to know the inner mathematical workings of the system but are more interested in the explanation itself. 

\vspace{10cm}
\subsection*{Healthcare}

XAI in the domain of healthcare can be viewed from a number of perspectives: technological, legal, medical and patient. Firstly, from a purely technological perspective, explainability will not only help improve model performance itself but also help sanity check the AI system for undesirable inference patterns. From a legal perspective, data acquisition, storage, transfer, processing and analysis need to comply with privacy requirements and other legal requirements. Explanations in particular are important from a legal perspective to establish accountability in XAI systems for tasks ranging from prediction analysis to AI based medical products. In fact, A recent FDA publication (\cite{FDA}) about artificial intelligence in medical devices states that an “Appropriate level of transparency (clarity) of the output and the algorithm aimed at users” is required. As a result, for AI to be effectively integrated into the medical industry, explanations must be generated for the decisions made by the AI systems deployed. From a medical perspective, explainable AI facilitates trust in the proposed AI system by the medical professional. The explanations provided would be of two levels. The first level would allow a medical professional to understand how the system makes a decision at a high level, where the explanations are generalized and do not refer to a particular case. The second level explains which features and variations in particular led to an individual decision for a particular case. These explanations would be helpful in resolving disagreements between the AI system and the medical professionals. From the patient’s perspective, explainability is key in establishing trust and a patient-centric approach to medicine. By providing patients with explanations and reasoning, they become more knowledgeable and informed of the decisions made regarding their health.

As a result, the audiences for XAI systems in the domain of healthcare include AI practitioners, regulatory bodies, medical professionals and patients. The goals of the XAI system vary for each of these audiences. For the AI practitioners the goal is to improve the performance of the AI model. For the regulatory bodies, establishing transparency and accountability is most relevant. Medical practitioners need to verify the correctness of the AI system and patients may want to understand the reasoning of the AI model in the context of their own health.

In the following subsections, we explore XAI systems that explain deep learning approaches for medical image analysis. Specifically, attribution-based methods, non-attribution-based, perturbation-based methods, backpropagation-based methods, intrinsically explainable methods, and attention-based methods will be discussed.

\subsubsection*{(1) Attribution-based Methods}

Attribution based methods refer to methods of explanation that assign an attribution value (represents contribution or relevance to the final output) to each input feature. In this section, we will discuss two types of attribution based methods (perturbation-based methods and backpropagation based methods) with relevant applications in the medical domain. 

\vspace{3mm}

\emph{\textbf{Perturbation-based Methods}}

\vspace{2mm}

Removing, masking, or modifying certain input features, and running the forward pass (output computation), and measuring the difference from the original output. Those input features which when masked, change the final outcome significantly are considered to be important features with a high attribution value. This method of explanation is model agnostic since the change is made to the data input and not the model itself. A popular perturbation based method is introducing occlusion (blocking out areas of an image) to input image data. An application of this method shall be discussed later.

\vspace{3mm}

\emph{\textbf{Backpropagation-based Methods}}

\vspace{2mm}

In these methods, the attribution value of a feature depends on the gradients that are calculated for that feature in the forward and backward pass through the network. Some popular backpropagation based methods include Gradient and DeConvNet. The gradient method, a fairly simple approach, sets the attribution value of each feature to the gradient of the output neuron with respect to the input neuron for that feature. DeConvNet is typically used for networks that use rectified linear unit  (ReLU) activations and applies the ReLU function to these gradients so that the attribution value represents the features learnt by the neurons.

\vspace{3mm}

\emph{\textbf{Application of an attribution-based Method}}

\vspace{2mm}

In a recent study by \cite{Singh2020}, explainability models (including Occlusion, Gradient and DeConvNet) were compared in the scenario of detecting choroidal neovascularization (CNV), diabetic macular edema (DME), and drusens from optical coherence tomography (OCT) scans. The image in Fig. 4.7 shows the results of the explanations and visualizations produced. The red regions depict positive influence and the blue regions depict negative influence on the network output.

\begin{figure}[H]
    \centering
    \includegraphics[width=8cm]{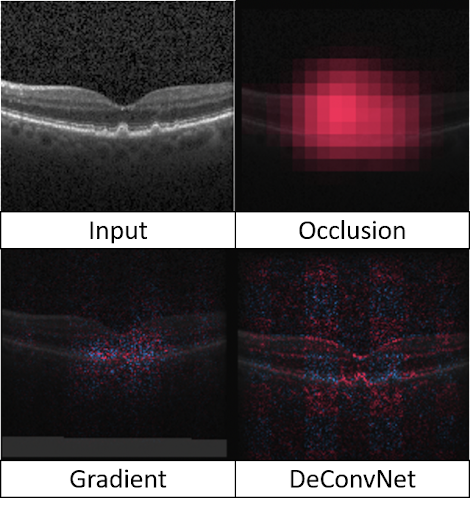}
    \caption{The results of the explanations for the attribution based methods}%
    \label{figure:atr}
\end{figure}

As can be seen in Figure \ref{figure:atr}, occlusion is not able to distinguish between the negative and positive outputs and also produces an explanation for gross features in the input. The backpropagation methods discussed here seem to produce more refined explanations that highlight important features and describe their positive or negative influence on the network output. In addition, perturbation based methods require the network to be run for each input feature that needs to be masked while backpropagation based methods require significantly less number of forward and backward passes through the network to retrieve the attribution values for the input features. 

\subsubsection*{(2) Non Attribution-based Methods}

Non attribution-based methods generate explanations by developing a suitable methodology for a given problem rather than performing an analysis on the input features (as seen in the previous section).

\vspace{3mm}

\emph{\textbf{Attention Methods}}

\vspace{2mm}

\begin{figure}[H]
    \centering
    \includegraphics[width=\linewidth]{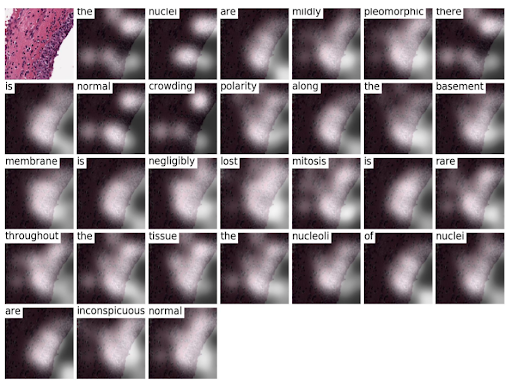}
    \caption{Visualization that depicts which part of the image predicts which word in the explanation}%
    \label{figure:atn_words}
\end{figure}

Attention-based methods identify which parts of an image influence the decision of the AI most significantly. This is different from attribution-based methods because these methods do not analyze the impact of the input data on the output explicitly. Instead, they exploit the attention weights already encoded into the model during training to derive the explanation. An example of this method can be seen in MDNet (\cite{zhang2017mdnet}). MDNet uses attention maps to explain the mapping between medical images and AI generated diagnostic reports. Figure \ref{figure:atn_words} an example of this, that depicts which parts of the input image was most significant in predicting which word in the output sentence. 

\vspace{3mm}

\emph{\textbf{Concept vectors}}

\vspace{2mm}

\begin{figure}[H]
    \centering
    \includegraphics[width=\linewidth]{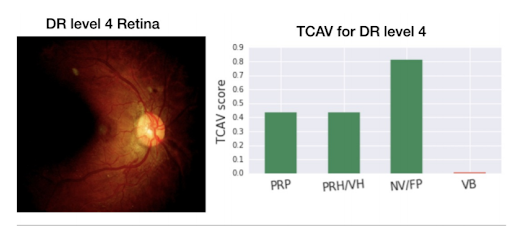}
    \caption{TCAV analysis that quantifies the importance of user-defined concepts}%
    \label{figure:tcav}
\end{figure}

Concept vectors provide an interpretation of a neural net’s internal state in terms of human-friendly concepts. The TCAV method (Testing with Concept Activation Vectors) by \cite{kim2018interpretability} uses directional derivatives to quantify the degree to which a user-defined concept is important to a classification result–for example, how sensitive a prediction of zebra is to the presence of stripes. This method was applied to the task of predicting diabetic retinopathy (DR) from retinal fundus images. The results can be seen in Figure \ref{figure:tcav} that analyzes the importance the model gives to concepts such as panretinal laser scars (PRP), preretinal hemorrhage and vitreous hemorrhage (PRH/VH), new vessels and fibrous proliferation (NV/FP) and a non diagnostic feature such as venous beading (VB). The level refers to the level of DR identified (ranges from 0-no DR to 4-proliferative).

\vspace{3mm}

\emph{\textbf{Expert Knowledge}}

\vspace{2mm}

In these methods, domain-specific knowledge is used to craft rules for prediction and explanation. \cite{pisov2019incorporating} integrate task specific knowledge into CNNs for brain midline shift detection. The expert knowledge integrated into this system are structural features of a brain midline that improves the accuracy of midline localization. They claim that since midline localization, a major part of the process of midline shift detection, is guided by expert rules, that the AI system as a whole becomes more interpretable because we have more insight into the driving factors behind the decision of the AI. However, no explicit explanations are discussed in this paper. 

\vspace{3mm}

\emph{\textbf{Textual Justification}}

\vspace{2mm}

\begin{figure}[H]
    \centering
    \includegraphics[width=\linewidth]{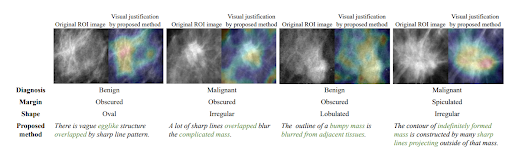}
    \caption{Textual explanations generated to classify the masses as malignant or benign}%
    \label{figure:text_jus}
\end{figure}

Textual justification methods generate explanations in the form of natural language text that makes these explanations understandable by both experts (medical professionals) and general audiences. In \cite{lee2019generation}, a justification model aims to explain the diagnostic decision of a deep network for the classification of breast masses as malignant or benign. The justification model takes as input the encoding of the input image as well as the diagnosis prediction of the model and outputs two kinds of justifications, a visual and a textual justification. The visual justification is obtained using a heatmap of the attention vector used in the prediction network and the textual explanations are generated from an LSTM network. The results of this XAI system can be seen in Figure \ref{figure:text_jus}.

\subsection*{Manufacturing}

AI models are widely used in the manufacturing industry and companies often depend on these models for the smooth running of operations and minimizing downtime. These systems are extremely crucial since they may be part of a long pipeline of operations and an error at a single point often shuts down the whole pipeline till it is resolved. Hence, explanations for these AI models are highly desirable for monitoring and auditing purposes. The primary audience for these explanations would be field experts who need to trust the AI model before incorporating it into a system. 

AI is used in various facets of manufacturing such as fault diagnosis, predictive maintenance, manufacturing cost estimation, machining feature visualization and so on. These tasks fall into three categories: tasks that can be done by humans but take a long time (fault diagnosis), tasks that can be done by humans with mediocre results (machining feature visualization), tasks that cannot be done without the assistance of AI (predictive maintenance). All of these tasks require AI with different levels of explanation that would justify the use of the AI model in a particular setting.

The domain of manufacturing is different from the previously discussed domains as the benefactors of the AI technology are usually not end users and hence, explanations of AI models do not need to be catered towards that audience. For several tasks, AI is the only solution available (or is significantly faster and cheaper than the previous solution) and a minimal explanation/transparency is usually enough to convince the industry to adopt the AI solution. A more elaborate explanation may not be necessary but is definitely desirable for monitoring and auditing the AI model to further improve performance. Therefore, we should aim for explanations that are beyond satisfactory and can be used to improve the AI system.

\begin{figure}[H]
    \centering
    \includegraphics[width=\linewidth]{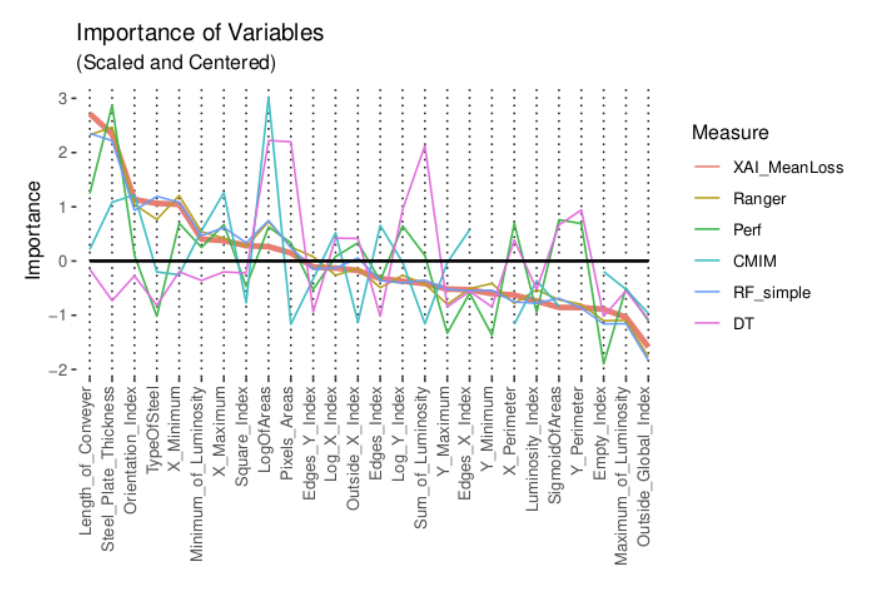}
    \caption{Contribution of each variable to the model decision}%
    \label{figure:man_imp}
\end{figure}

One such important task is fault diagnosis and prediction. Fault diagnosis is inspecting a material and identifying faults in it. In \cite{kharal2020explainable}, they take the material to be steel and the various bumps, dirtiness, scratches(different types), stains and so on. Fault prediction is understanding how these material flaws affect the lifespan of the material as a conveyer belt. We use an optimized random forest to understand these impacts. To understand how different features of the steel can affect the lifespan differently, we perform insight harvesting using different post hoc explanation techniques. First, we can see the relative importance of the features and how they contribute to the possibility of different failures. As observed from Figure \ref{figure:man_imp}, the length of the conveyer and thickness of the steel plate are the largest determiners of fault prediction for a steel conveyor. 

\begin{figure}[H]
    \centering
    \includegraphics[width=\linewidth]{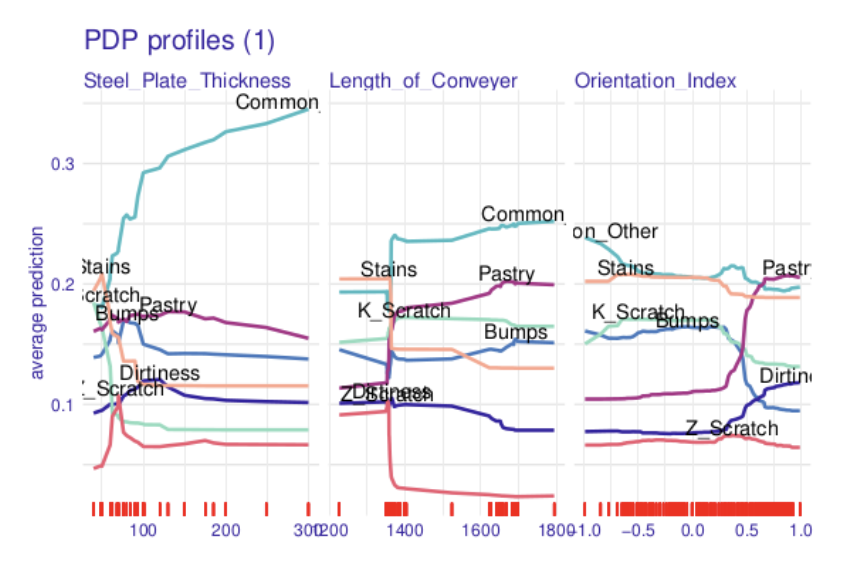}
    \caption{PDP profiles depicting how the variable affect the prediction}%
    \label{figure:pdp}
\end{figure}

These properties of the steel heavily influence the imperfections that inevitably occur in the creation process. To understand how they are related, we use partial dependence plots. Partial Dependence (PD) Plot describes how certain sets of variables affect an average prediction. For example, a thicker steel plate is less likely to have stains when compared to thinner steel plates as shown in Figure \ref{figure:pdp}. This method assumes that all imperfections are independent of each other. The red line at the bottom demonstrates confidence based on how much data is backing up the claim.

So, we first explain the optimized random forest used for fault prediction in steel conveyors and then analyze how surface imperfections ultimately affect the lifespan of the conveyor.

\section{Explanation Engineering}

Up to this point in the paper, we have seen that approaches to explainability can vary depending on the application domain. Even within a single domain, there are a variety of XAI techniques to choose from. Some of these methods are built alongside the machine learning model, while others can be used to generate explanations for models that are already trained. Some methods are specific to certain types of architectures (such as neural networks), while others are model agnostic (do not depend on the choice of model architecture). This raises some important design questions such as which XAI technique (among a few possible ones) is appropriate for the given task? In what format should the resulting explanation be? How can the quality of a model’s explainability be evaluated? It is important to answer such questions early on in a machine learning project since improper planning could greatly diminish the quality (and possibility) of the resulting explanations. Additionally, the choices of XAI techniques available at different stages of the design process vary (pre- vs post-model techniques), making it all the more necessary to think about one’s approach to explainability right from the start. 

We, therefore, propose the need for a formal framework for incorporating explainability into the typical machine learning workflow; we use the term “explanation engineering” (analogous to “requirements engineering”) as a name for this discipline. Based on a review of prior work in the field of XAI, we propose a 6-stage approach to develop an explanation system around a machine learning setup. For each stage, we suggest a minimal set of questions/concerns that need to be addressed; these are non-exhaustive, and should be used as a starting point to think about the requirements at each step of the design process.

\begin{figure}[H]
    \centering
    \includegraphics[width=\linewidth]{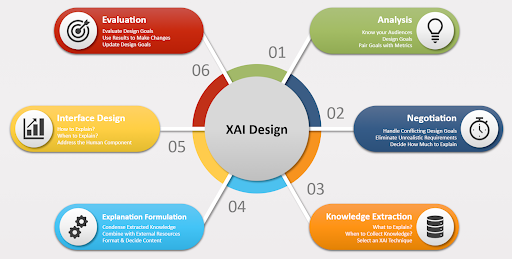}
    \caption{6-stage approach for XAI design}%
    \label{figure:xai_approach}
\end{figure}

\subsection*{Stage I: Analysis}

Our first step, as with most engineering approaches, is to analyze our explanation system’s requirements. This is important because it provides a way to ensure that the system is actually serving its intended purpose while also providing insights into changes that can be made to the system such that it can operate more effectively. This section presents a systematic approach to detailing and analyzing such requirements, specifically for the task of generating explanations for a machine learning system.

\subsubsection*{Know Your Audiences}

While ML systems are typically designed for a specific audience, explanations may be needed by other interested parties such as government regulators, insurance companies, among others, for various different reasons. Each of these audiences may have very different needs and, accordingly, it is important to start the explanation generation process by figuring out exactly who the audiences are for the given application. In the Table \ref{table:audiences}, we provide a list of the common choices of audience for most ML projects.

\begin{table}[H] 
  \caption{Description of audiences for explanations}
  \label{table:audiences}
  \raggedleft
  \begin{tabular}{|p{3.5cm}|p{9cm}|}
    \toprule
    \cmidrule(r){1-2}
    Audience & Description \\
    & \\
    \midrule
    End Users & The recipients of the machine learning system’s decisions. \\
    \midrule
    System Developers & Explanations are always helpful for the developers of the system so that they can improve predictions and debug errors. \\
    \midrule
    Researchers / Domain \newline Experts & Knowledge seekers interested in adopting ideas from the models. \\
    \midrule
    Government Regulators & If the ML system is adopted in public services, they need to pass certain tests as the risk of mistakes may be high. \\
    \midrule
    Insurance Companies / Financers & When a project needs to be financed through external sources, these third parties may benefit from explanations to judge the performance (and possibility of errors) of a system and determine if they want to invest in it. \\
    \bottomrule
  \end{tabular}
\end{table}

\subsubsection*{Design Goals}

In prior literature on XAI, there have been a few attempts to create a single, concrete definition for what an “explanation” is. Though these definitions do have some merit, they are generally at too high a level and do not capture the requirements of an explanation for every possible domain. Furthermore, there is no general consensus on any particular definition of an explanation. In light of this, we propose that a better approach would be to conceptualize explanations for a certain application as a list of design goals. Doing so, we can consider each of our audiences individually and itemize requirements for each of them.

In \cite{mohseni2020multidisciplinary}, the authors surveyed past literature and have put together a list of typical XAI design goals for three main audiences: AI Novices (which typically includes most end users), Data Experts and AI Experts. \cite{ribera_inproceedings} also advocate for a user-centric approach to XAI systems, and have proposed goals for similar user groups. Building on their work, we provide potential user needs for the audiences mentioned in the previous section in Table \ref{table:goals}.

\begin{table}[H] 
  \caption{Design goals for the different audiences}
  \label{table:goals}
  \raggedleft
  \begin{tabular}{|p{3cm}|p{3cm}|p{6cm}|}
    \toprule
    \cmidrule(r){1-3}
     & Design Goal & Description \\
     & & \\
     \midrule
    End Users & Transparency & User's understand how the system works. \\\cline{2-3}
        & Trust & Users can trust and rely on the system's decisions. \\\cline{2-3}
        & Bias Mitigation & Decisions made by the system are not biased. \\\cline{2-3}
        & Privacy Awareness & User's can assess their data privacy. \\\cline{2-3}
        & Curiosity & Users can learn more about the system. Increase interest. \\\cline{2-3}
        & Safety & Protect users from mistakes made by the system. \\\cline{2-3}
        & Interactivity \& Personalization & Users can ask for clarifications, provide feedback to personalize explanations, having multiple degrees of explanations. \\
    \midrule
    Researchers / \newline Domain Experts & Model Visualization \& Inspection & Analyze models, inspect for bias and deal with failure cases. \\\cline{2-3}
    & Model Tuning \& Selection & Use explanations to optimize parameters and improve model performance \\\cline{2-3}
    & Adoption & Develop new technology based on ideas from the system \\\cline{2-3}
    & Enlightenment & Use patterns identified by ML systems to further human knowledge.\\
    \midrule
    System Developers & Verifiability & Ensure that the system is performing the intended task \\\cline{2-3}
    & Improvement & Make changes to the system by understanding what it is doing from explanations. \\\cline{2-3}
    & Debugging & Explanations can provide clues to mistakes in the system, allowing developers to isolate the erroneous component(s). \\\cline{2-3}
    & Predictability & If users understand explanations, they may be able to predict the system's decisions in other scenarios. \\\cline{2-3}
    & Bias/Fairness Management & Detect bias in data, use of protected attributes in decisions. \\
    \midrule
    Government Regulators / Authorities & Compliance & Ensure that system's meet the minimum required standards. \\\cline{2-3}
    & Accountability & Explanations shed light on which party is responsible for errors made by the system. \\
    \midrule
    Insurance \newline Companies \newline / Financers & Evaluation & Explanations allow companies to judge a system and decide if they should invest/insure the developers. \\
    \bottomrule
  \end{tabular}
\end{table}

Another thing to keep in mind when thinking about design goals for an application is that the list does not need to be set in stone; we can always start with a few basic ones and add/remove goals as we see fit throughout the project cycle. This flexibility makes it possible to set a well-defined target for our system while also allowing us to address additional audiences and requirements that may arise as our environment changes. 

\subsubsection*{Pair Goals with Metrics}

Having a set of design goals is a good start, but it isn’t effective without a way to measure our system’s performance with respect to those goals. Accordingly, we need to think of one or more suitable metrics for each requirement while also considering the possible drawbacks/biases in the methods we select.  

In the prior subsection, we have listed popular design goals one may consider when working on a machine learning project. In Table \ref{table:evaluate_goals} we now introduce a few classes of strategies to evaluate these goals which have been assimilated from prior literature. 

\begin{table}[H] 
  \caption{Strategies to evaluate the goals of an XAI system}
  \label{table:evaluate_goals}
  \raggedleft
  \begin{tabular}{|p{2.5cm}|p{10cm}|}
    \toprule
    \cmidrule(r){1-2}
    Mental Models & Useful to evaluate whether users understand how a system works. \newline May be tested by assessing users’ ability to: \newline
    \begin{itemize}
        \item Predict model’s behavior on new inputs
        \item Identify cases where the model may perform poorly
        \item Understand how data is used to make decisions
    \end{itemize} \\\cline{2-2}
    & Design Goals: Transparency, Bias Mitigation, Privacy Awareness, Curiosity, Predictability \\
    \midrule
    User Satisfaction \newline / Usefulness & Evaluate whether users can trust and rely on the system. \newline May be tested by assessing: \newline
    \begin{itemize}
        \item Number of active users/populations
        \item The level of trust/reliance shown by users of the system
        \item Whether the system’s decisions benefit the user
    \end{itemize} \\\cline{2-2}
    & Design Goals: Trust, Curiosity, Evaluation \\
    \midrule
    Performance & Evaluate quantifiable properties of the system. \newline May be tested by assessing: \newline
    \begin{itemize}
        \item Increased latency from explanation generation
        \item Confidence scores for system’s decisions
        \item The system’s ability to protect users from errors
    \end{itemize} \\\cline{2-2}
    & Design Goals: Debugging, Improvement, Safety, Evaluation \\
    \midrule
    Fidelity & Judging whether explanations reliably explain what the system does. \newline May be tested by assessing: \newline
    \begin{itemize}
        \item Predictability of the system’s decisions/behavior
        \item Expected failure on tasks beyond system’s limits
        \item Generalizability of explanations to similar inputs
        \item Ability to identify and isolate sources of errors in the system
    \end{itemize} \\\cline{2-2}
    & Design Goals: Debugging, Verifiability, Compliance, Accountability, Evaluation \\
    \bottomrule
  \end{tabular}
\end{table}

\subsection*{Stage II: Negotiation}

Once we have defined the goals of our system, we must next consider any dependencies between them and resolve any conflicts that may arise. In this section, we address some of such compromises that need to be made when designing explanations for a machine learning system. 

\subsubsection*{Handle Conflicting Design Goals}

It may not be possible to achieve every single design goal for all our audiences simultaneously. Additionally, we must also consider objectives set for the machine learning system (such as model accuracy, latency, and so on), which makes this task all the more challenging. In Table \ref{table:competing_goals}, we provide a few examples of the type of conflicts one may encounter in typical machine learning projects.

\begin{table}[H] 
  \caption{Competing goals for XAI design}
  \label{table:competing_goals}
  \centering
  \begin{tabular}{|P{12cm}|}
    \toprule
    Transparency vs Performance \\
    \midrule
    This is the most common trade-off ML system designers need to make. While there are certainly simpler models such as decision trees and naive bayes, the performance achieved by the system may not be satisfactory. We may, therefore, be forced to use a more complex architecture such as deep neural networks, though these models are notorious for being inherently difficult to explain. Accordingly, we need to determine the right balance between model complexity (which directly affects transparency) and system performance. \\
    \midrule
    Quality of Explanations vs System Latency \\
    \midrule
    Many applications require the use of complex ML architectures. As such, systems designed to generate explanations for the predictions of such systems may also be complex. If explanations are incorporated into ML products, we would have an increased latency, possibly even beyond the actual prediction latency itself. Reducing the quality of explanations by using a simpler technique could reduce this latency, though the resulting explanations may not be good enough. Once again, we must trade-off these two design goals depending on what's best for our application.  \\
    \bottomrule
  \end{tabular}
\end{table}

Once we have identified possible conflicts among our design goals, we must then think of a suitable strategy to handle them. One approach is to prioritize audiences (as well as the model objectives) and determine which goals are most important to each. Then, in the event of a conflict, we must discard the design goal with the lower priority. A second, and likely a more useful, strategy is to find a satisfactory compromise between the two objectives. This may not always be an option but, depending on the application, it may be better than abandoning one of the design goals altogether.

\subsubsection*{Eliminate Unrealistic Requirements}

Sometimes, the design goals we come up with may not be practical. For example, let's consider that we use a decision tree with a moderate depth to implement a classification task. Since the model itself is simple and inherently explainable, we would not expect to be able to provide the user with multiple explanations of varying complexities. Accordingly, this cannot be one of our design goals. Simply having too many design goals may also be unrealistic. Trying to satisfy all of them simultaneously may require more effort and time than the developers can provide. We must, therefore, consider which of the design goals are actually important and which of them are beyond what the system designers can achieve.

\subsubsection*{Decide How Much to Explain}

In a medical diagnosis system, explanations provided to doctors may contain highly technical vocabulary that patients, insurance companies or government regulators may not fully comprehend. As such, explanations need to be tailored for each audience specifically. In addition to the technicality of explanations, we may also determine the minimum level of explanation needed to satisfy an audience. 

Consider a ML system that predicts the price for a house. One way to explain a prediction for a certain house is by finding an example of a house in our training set that is extremely similar to the one in question, but differs in a few attributes. Our system can then justify its predictions by demonstrating how it adjusted for the differences in those attributes (for example, if the one in the training set had 2 rooms, but the new one has 3, we would expect a reasonably higher price).

Another way to explain the same prediction is by analyzing the weights assigned to different inputs to the model and how they are combined to derive the output. Doing so, we can provide the user with an equation that approximates the predicted price of the house.

While both these explanations may satisfy a certain audience, the second one involves more effort. Additionally, while it may be possible to derive an equation for simple regression problems with only a few input variables, it would not work for more complex regression models or neural networks. Accordingly, while we do care that our explanations satisfy our audiences, it is necessary to not go overboard in coming up with the perfect explanation but rather to find the right level of explanation that is achievable with the least use of resources.

\subsection*{Stage III: Knowledge Extraction}

This is where Explanation Engineering connects with the rest of the machine learning workflow. At this stage, our goal is to extract as much useful information from the machine learning system as possible, so that this knowledge can be used to explain the behavior/prediction of the system. This could include the most-relevant features from a model (obtained using a dimensionality reduction technique), heatmaps (for image-related tasks), and so on.  

Since the type of information available and the method to extract this information depend on the machine learning architecture, this would be a good point to select an appropriate model (or models) suitable for the user objectives (if one does not already exist). Conversely, the choice of machine learning model and the explainable technique may be traded off to ascertain the ideal combination of the two, allowing for the extraction of most useful information.

In the following paragraphs, we discuss the most-relevant decisions that need to be made at this stage.

\subsubsection*{What to Explain?}

Our machine learning system can consist of multiple components. Some of these may perform rudimentary operations (file I/O) while others may contribute to the system’s decision-making capabilities (model weights, dataset, feature extraction component). Additionally, the behavior of some components may be self-evident (a simple feature extractor that computes daily average over data), while others, not so much (model weights). It is, therefore, necessary to think about which parts of the system need explaining, and to what extent. 

Most machine learning systems can be broken down into four main pieces: data, features, model, and data pipes. The explanation for any prediction may arise from one of these entities. It is, therefore, important to collect the right amount of information from each of these, such that we can explain the behavior of our system effectively. Below, we present a few questions for each of these components that one can consider to determine what information they need to collect to ensure that their system generates meaningful explanations.

\begin{enumerate}[label=\Alph*.]
    \item Data: \\
    \begin{enumerate}[label=\alph*.]
        \item Where did the data come from?
        \item What do the attributes mean?
    \end{enumerate}

    \item Features: \\
    \begin{enumerate}[label=\alph*.]
        \item How is feature engineering performed? 
        \item How are the features different from the data they were used to create? (i.e. what additional knowledge do these features provide?)
    \end{enumerate}

    \item Model(s): \\
    \begin{enumerate}[label=\alph*.]
        \item What are the inputs? (i.e. which attributes does the model have access to?
        \item Which of them are most relevant to a certain prediction?
        \item What is the model architecture? (neural network, decision tree, etc.) 
    \end{enumerate}

    \item Data Pipes: \\
    \begin{enumerate}[label=\alph*.]
        \item How are the outputs from different models (if more than one) aggregated?
        \item Any transformations to the outputs before making a prediction? (e.g. converting probability scores to class labels based on threshold)
    \end{enumerate}
    
\end{enumerate}

\subsubsection*{When to Collect Knowledge?}

In addition to the “what,” it is also important to think about “when” to generate explanations, since the explainable techniques available at any point in the project cycle may vary. Furthermore, each technique offers its own pros and cons, which may make them suitable (or unsuitable) for our explanation system’s objectives.  Table \ref{table:pre_post} provides potential reasons for deciding on a pre-ML vs a post-ML technique (using the building of the ML model as the frame of reference). In some cases, it may also be useful to collect different kinds of information during both stages and utilize them as required by the application.

\begin{table}[H] 
  \caption{Potential reasons for deciding on a pre-ML vs. a post-ML technique}
  \label{table:pre_post}
  \raggedleft
  \begin{tabular}{|p{2.5cm}|p{5cm}|p{5cm}|}
    \toprule
    \cmidrule(r){1-2}
    Time of Knowledge Extraction & Description & Potential Reasons for Selection \\
    \midrule
    Pre-ML & Incorporating explanation generation along with the model. These may include data interpretability methods such as PCA, clustering, etc. as well techniques designed specifically for an application. & Data/model weights may not be accessible later \vspace{5mm} \newline Explanations need to be generated in real time \\
    \midrule
    Post-ML & Post-hoc analyses with the aim to explain the decisions of an existing system. LIME, SHAPLEY and other similar methods are the most common ones that fall under this category. & The ML system is already built \vspace{5mm} \newline Typically include model-agnostic methods which may work well for systems that are not inherently explainable. \\
    \bottomrule
  \end{tabular}
\end{table}

Consider the case of self-driving cars. The actions selected by the ML system designed to operate the car are extremely time-sensitive. As such, the explanations themselves need to be real-time such that the driver can take control of the vehicle if the system makes a mistake. This limited time-frame doesn’t leave much room for a full-fledged post-hoc analysis of the system’s decisions in order to generate explanations. In such cases, incorporating an explanation component into the system (similar to a component that monitors the output of the system and checks for possible issues) may lead to quicker explanations suitable for the task at hand.

On the other hand, suppose we have a pre-trained ML model that is extremely good at detecting fraudulent credit-card transactions. Rather than building the system (with an explanation component) from scratch and attempting to obtain the same level of performance, it would be smarter to use the existing model and try to explain it (if possible). Additionally, we may also not have access to the data used to train the model which prevents the use of pre-ML techniques such as PCA.

\subsubsection*{Select an XAI Technique}

Having answered the two prior questions, the choice of XAI techniques available to us should be down to a manageable subset. It is, therefore, time to weigh the benefits of each technique and pick the one most suitable for the application.

Generally speaking, we have the choice between model-agnostic and model-specific techniques; selection of the method depends on the ML model (if one has already been built). Table \ref{table:sp_ag} lists some of the model-specific and model-agnostic techniques popularly used in the field of XAI.

\begin{table}[H] 
  \caption{Difference between model specific and model agnostic techniques}
  \label{table:sp_ag}
  \raggedleft
  \begin{tabular}{|p{3cm}|p{10cm}|}
    \toprule
    \cmidrule(r){1-2}
    & Applicable Models and Techniques \\
    \midrule
    Model-Specific & Linear Regression, Logistic Regression, GAM, GLM, Decision Trees, Decision Rules, RuleFit \\
    \midrule
    Model-Agnostic & LIME, Shapley Values, Partial Dependence Plot, Individual Conditional Expectation, Scoped Rules, SHAP \\
    \bottomrule
  \end{tabular}
\end{table}

If the model itself in use is easy to explain, which means that it is suitable for model-specific methods. We prefer using model-specific methods. However, due to the broad usage of different neural networks in different cases, model-agnostic methods will play a bigger role in industrial level usage of explanation engineering. 

Among model-agnostic techniques, we find LIME to be the most basic and general method which does not need lots of extra requirements such as the quality of datasets. So it may be preferable to select LIME as the first choice, and then use it as a reference against other XAI techniques. We can then look at things like the quality and size of the dataset we use to decide if we need to use more advanced, strict techniques like Shapley Values. As these model-agnostic techniques are not hard and expensive to implement, it is always suggested to use as more agnostic techniques as possible and then make a comparison among each other to get a better result.

\subsection*{Stage IV: Explanation Formulation}

Based on past literature, we have noticed that ML engineers often end the explanation generation process after the Knowledge Extraction step. While that may be acceptable for certain applications, in most cases, there is yet a bit more work to be done in order to obtain the desired level of explanation.

At this stage, the goal is to synthesize the knowledge extracted from our machine learning system and present it in a form that best achieves our explanation criteria.  

\subsubsection*{Condense Extracted Knowledge}

Depending on the application and the complexity of our machine learning system, we may have multiple types of information (pertaining to data, features, models and data pipes). We must now consider how these relate to each other. For example, we may use a dimensionality reduction technique to ascertain the top 3 features that contribute to the model’s output. However, we also need to know how these features were generated from the original data. Taking this a step further, it may also help to know where the data itself came from or how it was created, as that too may explain some of our system’s behavior. Accordingly, we need to reason among these types of knowledge and combine them (possibly in the form of a causal chain of events from the data to the model output) to achieve the full picture on what the machine learning system is doing.

\subsubsection*{Combine with External Resources}

In some cases, the knowledge extracted from the machine learning system may not be enough by itself. Consider the example of a ML model designed to diagnose a person’s symptoms. Using a dimensionality reduction technique, we may determine that temperature and heart rate were the two main features that led to the model’s prediction. However, this does not really explain how or why those symptoms are related to the predicted illness. In order to be able to obtain these kinds of explanations, we would require access to extensive knowledge about the human body. Additionally, we may also need a separate component to relate the information in such knowledge bases to the input features (here, symptoms) of the system. This once again enforces the need to select realistic goals for our explanation system, and to ensure that it has access to sufficient information needed to generate the desired explanations.

\subsubsection*{Format \& Decide Content}

The final step at this stage is to formulate the explanation. Once again, we need to consider each of our audiences one at a time. Our main goal here is to map the knowledge extracted from our system onto the user's mental model. To do this, we must first consider the right format to deliver our explanations. This could be text, images and so on. Quite recently, \cite{ijcai2018p865} even proposed a voice-enabled system where users can have a dialogue with the system that would then explain its decisions in a way the user understands.

In addition to the format of explanations, we must also consider its content. In stage II (Negotiation), we identified the level of explanation suitable for each of our audiences based on their level of understanding of the subject matter as well as the kind of explanations needed to satisfy them. Accordingly, we now need to generate explanations based on these criteria by using the knowledge extracted from the ML system or from supplementary resources. 

In Table \ref{table:explanations}, we list a few types of explanations that can be generated for ML systems; examples of the use of each type of explanation are also provided.

\begin{table}[H] 
  \caption{Types of explanations generated for ML systems}
  \label{table:explanations}
  \raggedleft
  \begin{tabular}{|p{3cm}|p{4.5cm}|p{4.5cm}|}
    \toprule
    \cmidrule(r){1-2}
    Feature Presence & Identifying the presence or absence of characteristic properties. & A sentiment analysis output may be explained by the phrases in the text that contributed most to the prediction. \\
    \midrule
    Explanation by Examples & Presenting examples from the training set that have the same prediction and demonstrating similarities. & A credit card fraud prediction may be explained by showing other examples of fraud. \\
    \midrule
    Counterfactuals & Presenting examples from the training set that belong to classes other than the one predicted and illustrating why the input does not belong to those classes. & If a cancer detection model predicts that the tissue in a certain pathology scan is cancerous, it could provide an example of a pathological scan that does not contain cancerous tissue to present the difference. \\
    \bottomrule
  \end{tabular}
\end{table}

\subsection*{Stage V: Interface Design}

Though our system may generate explanations at this point, we must also consider how we present these explanations to the end user. This section draws on ideas from Human-Computer Interaction and brings attention to a few main design considerations from the perspective of explanations. Some of the points in this section were inspired by the book by \cite{Hulten_2018}); readers interested in the design of machine learning systems may refer to this. 

\subsubsection*{How to Explain?}

Most of the time, users may not need an explanation; it’s only when something goes wrong that users want one. In such a scenario, it may be useful to wait for the user to ask for an explanation (the “wait and see” strategy). This could be implemented in the form of a push button placed in a readily-accessible location on the user’s screen. Alternatively, if the system has a low confidence in any of its predictions, it could also prompt the user (via a pop-up notification) to be cautious and consider the explanation along with the prediction to make the final call. While this kind of approach may work for typical users, data scientists or AI experts who wish to monitor and debug the system may require a constant output of explanations. This can be achieved by annotating the system’s output with the explanation. In Table \ref{table:approaches}, we list the drawbacks of each method, which may be useful in determining the right approach (or a hybrid one) for an application.

\begin{table}[H] 
  \caption{Approaches to Providing Explanations}
  \label{table:approaches}
  \raggedleft
  \begin{tabular}{|p{4cm}|p{4cm}|p{4cm}|}
    \toprule
    \cmidrule(r){1-3}
    Mode of Explanation & Pros & Cons \\ 
    \midrule
    User-Centric & Leaves control in the hands of users & Hard to collect telemetry \\ 
    \midrule
    Prompts & Usually easy to collect telemetry \vspace{5mm} \newline May help mitigate risk of errors & May fatigue users \\ 
    \midrule
    Annotation & Useful for steady and continuous output & Passive—users may not even notice \vspace{5mm} \newline Takes up space on interface \\
    \bottomrule
  \end{tabular}
\end{table}

\subsubsection*{When to Explain?}

If we decide to prompt the user with explanations or annotate them within the system’s interface, we must also consider the timing and frequency of the explanations. Should the explanations be provided right after the prediction, or is it better to wait for a bit before doing so (providing a sort of summary for the system’s actions)? And if explanations are provided, how long must the system wait before prompting/annotating another to ensure that the user isn’t fatigued? Such questions need to be carefully considered before integrating explanations with the system’s interface.

Consider the example of a self-driving car. If we want our system to be able to explain its actions, such as why it is making a turn, when is the right time to provide the explanation? If the explanation is provided before the car makes the turn, the user can take control of the wheel and override the car’s actions in the event that making the turn was the wrong choice (for e.g., if there was a pedestrian crossing the road that the car’s image recognition algorithm didn’t identify). While this is certainly beneficial in this case, it may not be appropriate for the system to provide an explanation every single time it makes a turn. 

It is also conceivable that users may need certain explanations after the fact. For example, if the self-driving car was involved in an accident, authorities may need to know what the system did (and why) at the time of the incident. In this scenario, our system would need to maintain a log of the actions it took, along with explanations for each decision, so that it is available to interested parties when they need it. 

\subsubsection*{Address the Human Component}

For certain applications, we may also need to consider the human component when designing the explanation interface. For example, we may decide that personalization is important for our application. In a machine learning system whose predictions are explained in multiple ways (perhaps based on the presupposed expertise of the audience), we may decide to leave the choice of explanation level up to the user, thereby allowing them to decide how much information they need before they can trust the system.

Another approach to make our system more user friendly is to use an interactive design for our interface. After we provide the explanation, we may allow users to ask follow-up questions (for e.g., if our system explains that given the user’s temperature and heart rate, they likely have a certain illness, the user may then ask the system whether the infection is viral or bacterial). Using the information collected during the Knowledge Extraction phase, our explanation system may attempt to provide further details about the machine learning models, thereby helping the users understand the explanation better.

Finally, it is also possible that the user may not find the explanations provided by our system to be satisfactory. In this scenario, the system may collect feedback from the user and update its algorithms such that it can tailor explanations to the user’s preferences. 

\subsection*{Stage VI: Evaluation}

Now that our explanation system is built, we must evaluate the quality of explanations generated by it, as well as its general performance. Some of these evaluation tasks may be performed earlier on in the project cycle (for e.g., by evaluating the knowledge extraction techniques, latency of explanation formulation, etc.), however, we have chosen to add this section after interface design since this is when our generated explanations can actually reach a wider audience, i.e. the end users of the system. This section covers the tasks one needs to perform at this stage.

\subsubsection*{Evaluate Design Goals}

Once our explanations are presented to users, we can evaluate our system by assessing each of our previously-defined metrics. To do this, we need to operationally define what data we collect (user feedback, model statistics, etc.), how we collect it (automatic monitoring/surveys), and finally, the computation we use to transform this telemetry into a numerical estimate of our system’s performance. In Table \ref{table:operational}, we briefly mention a few common evaluation strategies that may be useful for this task. Interested readers may also refer to the paper by \cite{hoffman2019metrics} which provides a more comprehensive list of evaluation metrics for XAI.

\begin{table}[H] 
  \caption{Operationalization strategies}
  \label{table:operational}
  \raggedleft
  \begin{tabular}{|p{3.5cm}|p{9cm}|}
    \toprule
    \cmidrule(r){1-2}
    Strategy &
    Description \\ \midrule
    Questionnaires &
    Set of predefined questions for users to rate explanations (perhaps on a likert scale) \\ \midrule
    Task Reflection &
    Ask users to explain in their own words how the system works \\ \midrule
    Automatic Telemetry Collection &
    Establish a pipeline to log system metrics such as latency, confidence scores, etc. \\ \midrule
    Comparative Testing &
    Use two or more explanation formulation techniques and provide the generated explanations to different sets of users \\ \midrule
    Simulated Experiments &
    Try out different inputs and evaluate explanations based on expected outputs \\
   \bottomrule
  \end{tabular}
\end{table}

\subsubsection*{Use Results to Make Changes}

If our system does not perform well on certain metrics, we may revisit prior stages of the design process and try out different techniques to improve results. 

For example, if our results indicate low explainer fidelity, we may consider trying out a different knowledge extraction technique to better model how the ML system works. Similarly, if latency is the issue, we can isolate the component that takes the longest and try a different approach, perhaps at the cost of other design goals such as explanation quality.

\subsubsection*{Update Design Goals}

When evaluating our system, we may also think up new goals or audiences that we may add to our current design. Conversely, we may also decide that certain goals aren't as necessary as we thought they were and may, therefore, eliminate them. 

This is especially important in large scale systems with thousands of users. Over time, user needs may change. Additionally, an increase in the number of users may put strain on the system which may require us to consider other design goals to handle such issues. This step is, therefore, crucial in ensuring that our system is able to adapt to changing times while remaining useful to our users.

\section{Conclusion}

This paper highlights the importance of incorporating explainability in AI systems. First, to prime our understanding of explainable AI, the meaning of the term ‘explanation’ is discussed in detail. Both the contexts in which an explanation is produced and the audience to which the explanation is presented are explored in terms of how these factors change what constitutes a satisfactory explanation. In addition, we identify desirable qualities of an explanation and their evaluation schemes.  With this in mind, we take a look at various XAI techniques used to produce explanations. In particular, we organize popular XAI techniques under different categories such as global or local and model specific or agnostic. Specific techniques that provide explanations to regression models are also reviewed. The application of XAI techniques including ones that provide explanations for deep learning models, are examined under four domains, namely the domains of finance, autonomous driving, healthcare and manufacturing. Within each of these domains, the need for explainability, the audiences of these explanations, and examples of applications and the quality of the explanation produced are discussed. This section brings to light that both the choice of XAI technique and its implementation in the context of the application domain is crucial to the overall quality and satisfiability of the explanation produced. With this insight, we propose a novel discipline of explanation engineering that aims to set guidelines to incorporate explainability into AI systems. The importance of explainability reinforces the need to engineer explanations into AI systems as part of fundamental design decisions and not as an after-thought. With explanation engineering, we provide a 6 stage framework to navigate the incorporation of explainability in a typical AI workflow. We understand that the points covered in explanation engineering are not exhaustive and can be detailed further with more time and deliberation. 

Going forward, we realize that a major challenge that bottlenecks the progress of XAI systems is the subjectivity of an audience’s satisfaction towards an explanation. This also makes the evaluation of explanations non-trivial because a “good” explanation for one person may not be a satisfactory explanation for another. As future work, a hierarchical structure to explanations can be explored that modifies the level of explanation provided based on audience feedback. In addition, the verifiability of explanations is a significant yet often overlooked problem with XAI systems. In the future, verification systems that validate the truthfulness of the explanation produced can also be analyzed and integrated into the explanation engineering workflow.

\bibliographystyle{plainnat}
\bibliography{ref}

\end{document}